\documentclass[runningheads]{llncs}

\usepackage[mobile]{eccv}

\usepackage{eccvabbrv}

\usepackage{graphicx}
\usepackage{booktabs}
\usepackage{multirow}
\usepackage{float}
\usepackage{longtable}

\usepackage{xcolor}

\usepackage{pifont}

\usepackage[accsupp]{axessibility}  

\usepackage{hyperref}

\usepackage{orcidlink}

\begin{document}

\title{Dynamic Inverse Rendering for Enhanced Material-Lighting Decomposition} 


\author{Raza Yunus\inst{1,3} \and
Benjamin Ummenhofer\inst{2} \and
Jan Eric Lenssen\inst{3} \and
Eddy Ilg\inst{1,\dagger}}

\authorrunning{R.~Yunus et al.}

\institute{University of Technology Nuremberg \and
Intel \\
\and
Max Planck Institute for Informatics, Saarland Informatics Campus \\
\vspace{1.75pt}
{Project Page: \href{https://razayunus.github.io/DIR}{razayunus.github.io/DIR}. $^\dagger$ Now at Google.}
}

\maketitle

\begin{center}
    \centering
    \captionsetup{type=figure}
    \includegraphics[width=\textwidth]{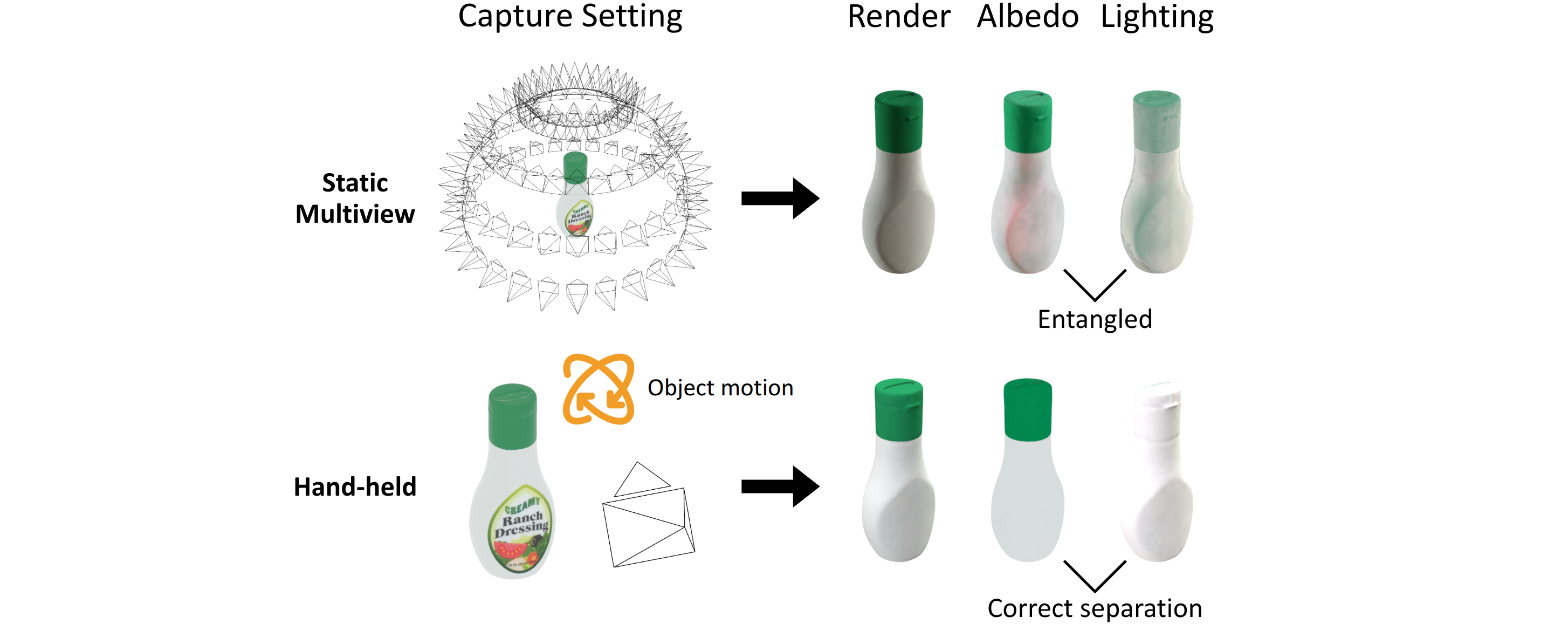}
    \captionof{figure}{In contrast to the static multiview setting, where albedo and the corresponding lighting often become entangled, we use the hand-held object capture setting in which diverse surface-light interactions impose strong constraints on the optimization and enable proper material-lighting disentanglement.}
    \label{fig:teaser}
\end{center}

\begin{abstract}
  Decomposing outgoing surface radiance into material and illumination during inverse rendering is essential for applications such as relighting and augmented reality, yet it is severely ill-posed since multiple combinations can result in the same observed colour. Capturing an object under multiple lighting conditions usually helps resolve this ambiguity as it constrains the optimization towards correct solutions. In this work, we explore the potential of reconstructing rigidly moving objects---which provides observations of diverse light-surface interactions---to resolve the material-lighting ambiguity in inverse rendering. For this purpose, we introduce a relightable approach that marries object tracking and reconstruction with inverse rendering for general rigidly moving objects. Our experimental analysis on synthetic data demonstrates that motion can be an advantage for disentangling material and lighting: the reconstructed material is significantly more accurate when the object is observed under rigid motion than when it is static. Moreover, results on RGB videos of real hand-held objects show that our pipeline preserves this advantage even under noisy real-world conditions.
\end{abstract}

\section{Introduction}
\label{sec:intro}

Recent advances in radiance fields have achieved remarkable success in reconstructing scenes with high-fidelity appearance~\cite{mildenhall2020nerf,kerbl20233d} and geometry~\cite{wang2021neus,huang20242d}. However, these methods typically bake in the appearance under fixed lighting conditions, making them unsuitable for scenarios where relighting is required, e.g. placing reconstructed objects in virtual environments. While inverse rendering approaches estimate material and lighting separately for this purpose~\cite{jin2023tensoir,gao2024relightable}, the decomposition achieved is often suboptimal due to the inherent ambiguity in inverse rendering. This ambiguity is traditionally resolved by capturing a static object under multiple lighting conditions~\cite{jin2023tensoir,li2025recap} or utilizing either hand-crafted~\cite{gao2024relightable} or data-driven~\cite{chen2024intrinsicanything} priors. However, such methods require extra sophistication, from extended lighting representations for capturing varying lighting conditions~\cite{fei2024vminer,jin2023tensoir} to lengthy training procedures for learning useful priors from (limited) 3D data, either for surface materials~\cite{chen2024intrinsicanything}, environment illumination~\cite{lyu2023diffusion} or relighting~\cite{poirier2024diffusion} directly.
For most capture settings, a single far-field light representation can provide a decent approximation for the environment illumination. In this work, we posit that if an object is observed and reconstructed while undergoing significant motion, then the evolving light interaction between the object and a single far-field light representation provides enough observations to significantly enhance the material and lighting decomposition.

Since the single far-field light assumption requires tracking the object under large pose changes to observe significantly different light interactions, such dynamic reconstruction, while providing stronger constraints for material-lighting ambiguity, also demands accurate tracking, and tracking errors may degrade reconstruction quality. Although category-level or instance-level CAD models have been utilized~\cite{wen2024foundationpose} to track a rigidly-moving object in novel sequences, such methods assume that a reconstructed model is available and do not generalize to new objects, thus limiting their application. The most suitable setting for our purposes is the hand-held reconstruction setting, where a person freely rotates an object in their hands and a video is captured from a static viewpoint. Such a setting provides full 360 coverage of an object while also providing the varying lighting conditions we need to improve the material-lighting decomposition.

In this paper, we propose a relightable reconstruction pipeline that allows us to utilize the surface-light interactions of general dynamic objects for improved material estimation. We employ a three-stage approach to ease the optimization challenges of joint object tracking, reconstruction and inverse rendering. In the first stage, we use progressive sequential optimization together with the NeuS representation~\cite{wang2021neus} to obtain the coarse geometry and initial poses of the dynamic object. We utilize 2D correspondences from state-of-the-art feature matchers~\cite{edstedt2025romav2,leroy2024grounding} to aid with this optimization. In the second stage, we introduce 3D Gaussians and optimize them jointly with the object poses to obtain fine-grained geometry and pose estimates. Given the reconstructed geometry, we augment the Gaussians with material properties in the final stage and employ physically-based rendering to optimize these materials across timesteps, enabling them to explain the observed surface appearance under changing illumination. We further leverage efficient Gaussian ray tracing~\cite{moenne20243dgrt} to model accurate occlusion for the dynamic object within our inverse rendering pipeline. Finally, to quantitatively evaluate the improvement in material decomposition enabled by our 4D inverse rendering pipeline, we introduce a synthetic dataset of common household objects~\cite{banerjee2025hot3d}, along with ground-truth materials and relit views. We also demonstrate the practical viability of our method qualitatively on real-world hand-held object sequences.

Overall, we make the following contributions:
\begin{itemize}
    \item We posit and empirically validate that observing an object under significant motion provides stronger constraints for material-lighting decomposition than a static capture.
    \item We propose a 4D inverse rendering framework that brings together object pose tracking and surface material estimation.
    \item We introduce a synthetic dataset of common household objects with ground-truth materials and relit views, enabling controlled evaluation of material-lighting decomposition quality in three common capture settings: static multiview, turntable and hand-held capture.
\end{itemize}

\section{Related Work}
\label{sec:related-work}

Since our work brings together inverse rendering and dynamic object reconstruction, we briefly review state-of-the-art representations in the respective fields while also placing our work in the landscape of methods that specifically target better material-illumination decomposition.

\noindent\textbf{3D Reconstruction and Inverse Rendering.}
Recent progress in inverse rendering has been tightly coupled with advances in differentiable 3D scene representations. NeRF~\cite{mildenhall2020nerf} and 3D Gaussian Splatting~\cite{kerbl20233d}-based representations have become the de facto standard for high-quality novel view synthesis and 3D reconstruction. Since they bake view-dependent appearance into the geometry in their original formulation, methods~\cite{zhang2021nerfactor,boss2021nerd,liang2024gs,gao2024relightable} quickly extended these representations with material properties and utilized physically-based rendering~\cite{kajiya1986rendering} to optimize these materials and environment illumination separately, allowing relighting. Improvements in the underlying representation, such as enhancing the underlying geometric fidelity of the reconstruction~\cite{wang2021neus,yariv2021volsdf,huang20242d,yu2024gsdf} and improving efficiency with hybrid data structures~\cite{mueller2022instant,chen2022tensorf}, were adopted by the inverse rendering methods~\cite{li2024tensosdf,zhu2025gaussian,engelhardt2024shinobi,jin2023tensoir,sun2025svg}, boosting their efficiency and accuracy. In conjunction with the representation, enhancements have been introduced in the physically-based inverse rendering pipeline, including more accurate sampling strategies~\cite{hasselgren2022nvdiffrecmc,gu2025tensoflow}, varied shading functions~\cite{dihlmann2024subsurface,jiang2024gaussianshader,liu2023nero}, better illumination modelling~\cite{yao2022neilf,zhang2023neilf++,wu2023nefii,gu2025irgs} and handling unconstrained image collections~\cite{kuang2022neroic,boss2022samurai}. The discrete nature of 3D-GS-based representations and the fast rendering have made them especially adept at capturing fine details and modelling computationally expensive global illumination effects---enabled by ray tracing of 3D Gaussians~\cite{moenne20243dgrt}---efficiently, both of which are crucial for accurate inverse rendering.  However, in terms of modelling general objects, these representations have only been used in the static setting so far. We extend them to the dynamic reconstruction setting.

\noindent\textbf{Material-Lighting Ambiguity.}
A central challenge for inverse rendering methods is to resolve the ambiguity between surface reflectance and illumination. 
Under a fixed-illumination setup and without proper constraints, some of the object colour tends to get baked into the environment map, appearing as tints and shifts in tone and intensity. Vision science research has shown that humans are also bad at disambiguating between material and environment illumination~\cite{sylvia2006material}. Methods usually try to resolve this ambiguity through hand-crafted~\cite{hasselgren2022nvdiffrecmc,gao2024relightable} or learned~\cite{chen2024intrinsicanything,lyu2023diffusion} priors, or capturing the object in a multi-illumination setting. The former strategy focuses on learning priors for either the surface material~\cite{chen2024intrinsicanything}, environment illumination~\cite{lyu2023diffusion} or relighting directly~\cite{alzayer2025genrelight,zhao2024illuminerf}, hoping that the learned prior from large-scale datasets will constrain the optimization enough for proper disambiguation. The latter uses augmentation strategies, such as flashlight-enabled extra views~\cite{fei2024vminer}, multi-environment captures~\cite{li2025recap} or modelling external occluders~\cite{verbin2024eclipse} to introduce additional lighting constraints to resolve ambiguity. Powerful, diffusion-based priors have also been utilized~\cite{tang2025rogr,poirier2024diffusion} to produce multi-light views from a single lighting condition, which are then utilized in the optimization. Instead of training on massive amounts of data or modelling complex lighting setups, our approach explores the orthogonal setting of dynamic object reconstruction, which provides abundant lighting conditions through a single RGB video and requires a simple far-field light representation to capture the gain in material-lighting decomposition.

\noindent\textbf{Hand-Held Object Reconstruction.}
Inverse rendering methods usually require camera poses as input and assume a static scene. For dynamic, rigid reconstruction, the object has an additional six degrees of freedom to move in the observed scene. Most object pose tracking methods assume that an object model is available, either category-level~\cite{wang2023deep,tian2022large,wang20206,fischer2025unified} or instance-level~\cite{wen2020se,wen2024foundationpose}, which allows them to track the object through relative pose estimation but limits their generalizability. For novel object pose tracking, methods usually rely on RGB-D input, combined with a NeRF~\cite{wen2023bundlesdf} or Gaussian Splatting~\cite{jin20256dope}, to reconstruct and track the object. Some methods specialize in hand-held object tracking~\cite{fan2024hold,chen2025hort}, relying on hand pose annotations and consistent hand-object interactions to track the object. OnlineSplatter~\cite{huang2025onlinesplatter} introduces transformer-based 3D Gaussian tracking for real-time object-centric online reconstruction. More closely aligned with our setting of reconstructing and tracking free-moving objects in a globally consistent manner, the recent method FMOV~\cite{shi2025fmov} relies on NeuS and progressive pose optimization for the task, which is enhanced by utilizing correspondences from LoFTR~\cite{sun2021loftr}, followed by a global refinement step. In contrast, our method uses the state-of-the-art matchers~\cite{edstedt2025romav2,leroy2024grounding} for more reliable feature matching and introduces 3D-GS for the global refinement step, which allows fine-grained pose refinement since 3D Gaussians are better at capturing details, resulting in greater robustness for the inverse rendering task.

\section{Motivation and Problem Setup}
\label{sec:motivation}

Our central thesis is that observing a dynamically moving object from a stationary camera yields a richer set of surface-light interactions than a moving camera capturing a static scene. Greater lighting diversity allows us to better condition the ill-posed inverse rendering problem, leading to better material estimates.

A common paradigm for capturing dynamic objects is the hand-held setup, where the user freely rotates a rigid object in front of a stationary camera. Let $\{I_t\}_{t=1}^T$ be a sequence of $T$ images observing a rigidly moving object with unknown per-frame poses $\{(R_t, t_t)\}_{t=1}^T$, with $(R_t,t_t) \in SE(3)$. For an observed image at time $t$, points $x$ on the canonical surface $S$ and their corresponding normals $\hat{n}$ are transformed according to the object pose as:
\begin{equation}
   x_t = R_t x + t_t; \;\;
   \hat{n}_t = \hat{n}(x_t) = R_t \hat{n}
\end{equation}
We assume that the environment does not change over time and---given the far-field illumination assumption---the environment light reaching a surface point $x$ on the object from direction $\hat{\omega}$ is given by:
\begin{equation}
    L_i(x, \hat{\omega}) = L_{direct}(\hat{\omega}) V(x, \hat{\omega}) + L_{indirect}(x, \hat{\omega})
    \label{eq:illumination}
\end{equation}
where $V \in [0, 1]$ represents the visibility of the environment from the given surface point $x$ in direction $\hat{\omega}$, $L_{direct}$ is the far-field illumination and $L_{indirect}$ captures the global illumination effects. Combined with the spatially varying BRDF $f$, which defines the material at surface point $x$, the outgoing radiance at that point for a view direction $\hat{\omega}_o$ at time $t$ is defined as:
\begin{equation}
    L_o(x, \hat{\omega}_o, t) = \int_{\Omega_t \subseteq \mathbb{S}^2} f(x_t, \hat{\omega}_{it}, \hat{\omega}_o) L_i(x_t, \hat{\omega}_{it}) (\hat{\omega}_{it} \cdot \hat{n}_t)_+ d\hat{\omega}_{it}
    \label{eq:rendering}
\end{equation}
where $\Omega_t$ is the spherical domain of integration and $()_+$ is the positive clamping function.

Given a monocular video of a rigidly moving object, our goal is to recover the scene parameters, namely the object geometry $\{x, \hat{n}\} \in S$,
material properties, i.e. albedo $a$ and roughness $r$, per-frame object poses $\{R_t, t_t\}_{t=1}^T$ and the environment illumination $L$. This is achieved by optimizing the following objective for each rendered pixel $u$ at time $t$:
\begin{equation}
    \operatorname*{argmin}_{x, \hat{n}, \{R, t\}_t, a, r, L} \sum_{t, u} | I_\text{rend}(u; x, \hat{n}, R_t, t_t, a, r, L) - I_t(u) |
    \label{eq:objective}
\end{equation}
where $I_\text{rend}(u, t)$ is rendered using a differentiable renderer and $I_{t}$ is the observed ground-truth image at time $t$.

From Equation~\ref{eq:rendering}, we can see that the canonical object surface points $x$ interact with the environment illumination $L$ across observations through the rotated surface normals $R_t \hat{n}$. Given high-frequency lighting and observations of the object undergoing significant pose changes, the same surface point is seen under numerous lighting conditions, introducing more constraints to the optimization in Equation~\ref{eq:objective} than the static-object multiview capture setting, where no such interactions are observed.

\section{Method}
\label{sec:method}

\begin{figure}[t]
    \centering
    \includegraphics[width=\textwidth]{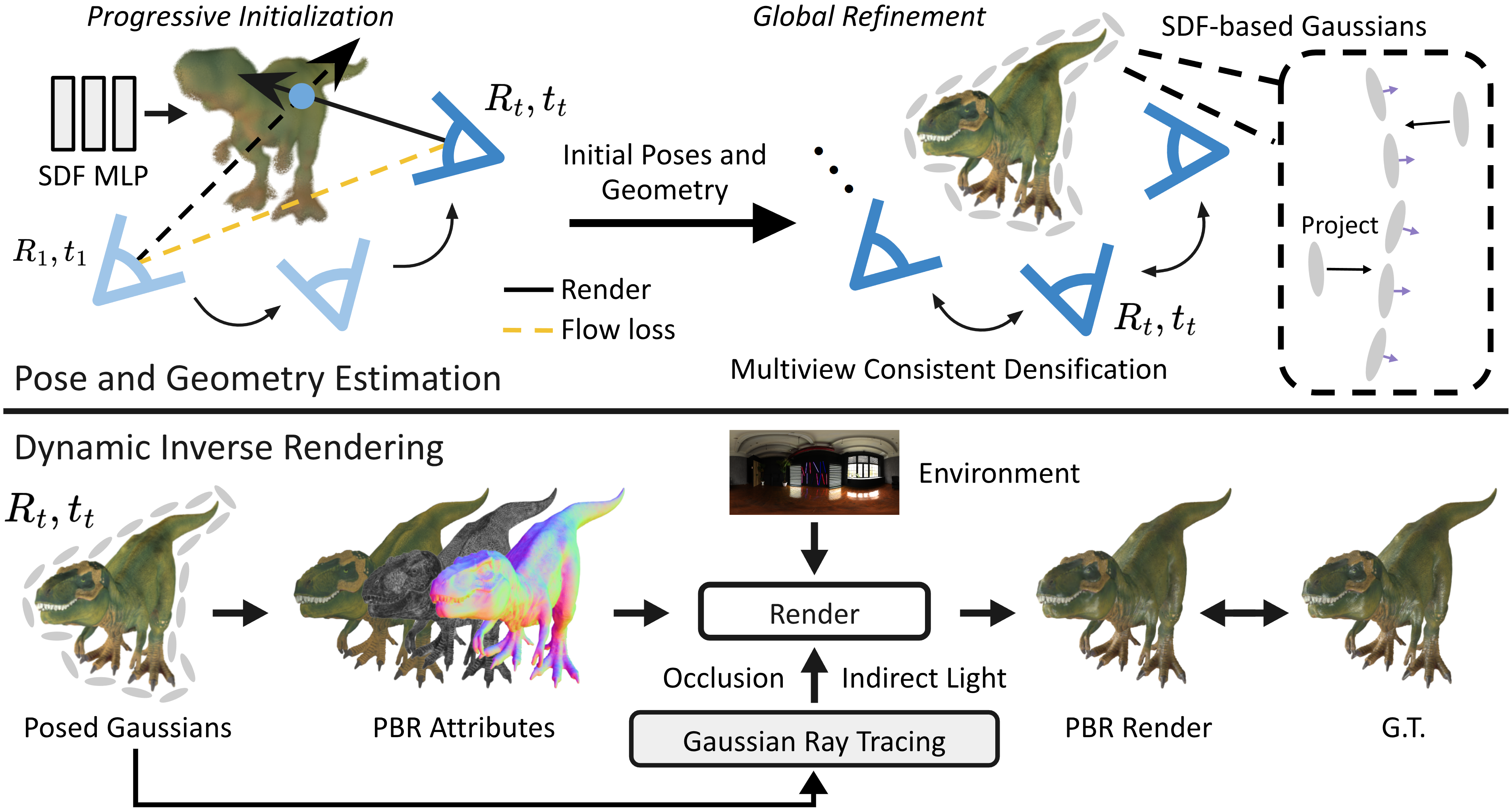}
        \caption{Overview of our three-stage method. We first perform progressive online optimization with a NeuS representation to estimate initial object poses and geometry. We then refine them with an SDF-based 3D Gaussian representation to capture fine-grained geometry. Finally, we add surface materials to the Gaussians and jointly optimize them with environment illumination to obtain accurate material-lighting decomposition from a moving object.}
    \label{fig:method}
\end{figure}

The objective in Equation~\ref{eq:objective} is severely ill-posed and estimating all unknown scene parameters jointly from the start is prone to getting stuck in suboptimal local minima. To tackle this problem, we introduce a multi-stage pipeline, as shown in Figure~\ref{fig:method}. First, we obtain the initial geometry and object poses through a progressive optimization scheme that uses a neural SDF representation for the object (Section \ref{sec:progressive}). Next, we switch to the 3D Gaussian representation and refine the initial estimates jointly across all frames to capture the fine details (Section \ref{sec:global}). Finally (Section \ref{sec:inverse_rendering}), we add materials to our representation and model the environment illumination, optimizing them jointly with the geometry and poses through physically-based rendering to accurately decompose material and lighting from observations of a moving object.

\subsection{Progressive Pose and Geometry Estimation}
\label{sec:progressive}

Estimating the pose and geometry is a classic ill-posed problem, and we need to optimize both iteratively to achieve robust solutions. We take inspiration from FMOV~\cite{shi2025fmov} to estimate object poses progressively with a neural SDF representation. The key insight here is to learn the geometry on a coarse level, to prevent the poses from overfitting to high-frequency noise, while still getting a sufficiently good initial estimate for our second-stage refinement.

\noindent\textbf{Neural SDF Representation.}
To facilitate progressive pose optimization, we adopt NeuS~\cite{wang2021neus,shi2025fmov} to initially represent the geometry, making use of the regularization properties of MLPs and the unit sphere geometric initialization of the SDF network, which keeps the surface intact under sparse observations. Given a point sample $x$ along the view direction $v$, the object's surface and appearance are modelled as:
\begin{equation}
    s(x), c(x, v) = F(x, v; \theta)
\end{equation}
where $F$ is an MLP network with parameters $\theta$, $c$ is the view-dependent colour and $s$ is the SDF value at point $x$. The rendered colour for a pixel $u$ is derived using the standard quadrature approximation of the volume rendering integral~\cite{mildenhall2020nerf}:
\begin{equation}
    I(u) = \sum_{k=1}^K w_k c_k
    \label{eq:compositing}
\end{equation}
where $K$ is the number of samples along the ray---kept low here to model coarse geometry---and colour $c_k$ is alpha-blended using blending weights $w_k$, composed of the transmittance and the volume density derived from the underlying SDF value~\cite{wang2021neus}.

\noindent\textbf{Per-Frame Pose Representation.}
Since the camera/object pose across the captured image sequence is unknown, we jointly optimize it with the geometry. The camera rays used to query the SDF MLP are transformed to each timestep $t$ using the pose $(R_t, t_t)$ predicted by a per-frame pose MLP. We also adopt the virtual camera system from FMOV~\cite{shi2025fmov}, which eliminates two degrees of freedom in translation by centering the object in each image using 2D masks. Although not physically accurate, this eases the pose optimization for noisy, real-world sequences.

\noindent\textbf{Sequential Optimization.}
We sequentially optimize the geometry and track the pose frame-by-frame---where each frame is initialized by the pose of the previous frame---to handle large pose changes and also reset the geometry once the pose has changed significantly, so that the degraded surface quality does not hinder further pose tracking~\cite{shi2025fmov}. We also use 2D feature correspondences~\cite{edstedt2025romav2,leroy2024grounding} in the neighbourhood of the current frame to constrain tracking by imposing cross-frame consistency; given that hand-held objects are often textureless and specular, the prior from a learned matcher helps regularize the tracking in such cases. The final loss at this stage is:
\begin{equation}
    \mathcal{L}_{progressive} = \mathcal{L}_{render} + \mathcal{L}_{eikonal} + \mathcal{L}_{mask} + \mathcal{L}_{match}
\end{equation}
where $\mathcal{L}_{render}$, $\mathcal{L}_{eikonal}$ and $\mathcal{L}_{mask}$ are the photometric, eikonal and mask losses from NeuS~\cite{wang2021neus} and $\mathcal{L}_{match}$ is the flow loss from FMOV~\cite{shi2025fmov}. At each step, only the pose MLPs of the current frame and the frame with matched correspondences are optimized.

\subsection{Global Pose and Geometry Refinement}
\label{sec:global}

Given the progressively optimized poses from the previous stage, we can now jointly refine them with the geometry at a finer level to obtain a more detailed reconstruction. Surface normals are also estimated for the geometry at this stage, and obtaining a good estimate is crucial for accurate material estimation later.

\noindent\textbf{3D Gaussian Representation.}
For the refinement stage, we introduce 3D Gaussians~\cite{kerbl20233d} as our geometry representation since they are good at capturing fine details. We integrate an SDF directly with the Gaussians, as introduced in DiscretizedSDF~\cite{zhu2025gaussian}, which helps stabilize the geometry during inverse rendering, especially for shinier objects. Each Gaussian is defined as an ellipsoid with a position $\mu$, a covariance matrix $\Sigma$ (parameterized as a quaternion rotation $q$ and a 3D scaling factor $s$), an SDF value $\sigma$ and view-dependent colour $c$ represented using spherical harmonics. The normal of a Gaussian is defined as the rotation axis corresponding to the smallest scale. The opacity is derived from the SDF as:
\begin{equation}
    O(\sigma) = \frac{4 e^{-\gamma \sigma}}{(1+e^{-\gamma\sigma})^2}
\end{equation}
with learnable global variance $\gamma$, while the response of the Gaussian $G$ is defined as:
\begin{equation}
    G(x) = \exp(-\frac{1}{2} (x - \mu)^T \Sigma^{-1} (x - \mu))
\end{equation}

During rendering, 3D Gaussians are projected to the 2D image plane and alpha composited using Equation~\ref{eq:compositing}, with the blending weights $w_k$ determined by the SDF-derived opacity $O(\sigma)$ and the response $G(u)$ for the projected 2D Gaussian at pixel $u$. Gaussians with backfacing normals are culled during rasterization, which avoids entanglement of front and back surfaces---always well-defined through the use of sphere initialization~\cite{zhu2025gaussian}---and helps solidify the SDF on the correct surface only~\cite{sun2025svg}. We also use geometry-aware rasterization~\cite{radegs} for rendering depth, which is determined directly by the intersection point of a ray with a Gaussian in 3D instead of using the Gaussian center, thus improving fidelity.

To render a Gaussian at time $t$, we first transform it using the object pose $(R_t, t_t)$. This results in transforming the position $\mu$ and rotation $q$ as:
\begin{equation}
    \mu_t = R_t \mu + t_t; \;\; q_t = Q(R_t)q
\end{equation}
where $Q(\cdot)$ converts a rotation matrix to a quaternion. Note that since the normal is derived from the Gaussian itself, it is already in the correct orientation when the transformed Gaussian is rendered. The spherical harmonics are evaluated equivalently by rotating the query direction with $R_t^T$ instead. Transforming the Gaussians rather than the cameras also allows the gradients to flow to the pose MLP for joint optimization~\cite{colmap-free-gs}.

\noindent\textbf{Global Pose Representation.}
For pose refinement, we use a global MLP with per-pose embeddings to learn deviations from the poses estimated during the first stage. Since the estimates from the previous stage were obtained using virtual cameras, an initialization for the full 6-DoF poses is computed from them using 3D-2D correspondences from the estimated mesh with the RANSAC EPnP algorithm~\cite{shi2025fmov}.

\noindent\textbf{Adaptive Density Control.}
We incorporate the multi-view consistent densification and pruning strategy introduced by FastGS~\cite{ren2026fastgs} into our pipeline. This keeps the number of Gaussians low and consolidates them at the correct locations, which improves the efficiency of the compute-intensive inverse rendering and 3D Gaussian ray tracing in the next stage.

\noindent\textbf{Optimization.}
At this stage, we globally optimize the geometry and all object poses using the following objective:
\begin{equation}
    \mathcal{L}_{global} = \mathcal{L}_{render} + \mathcal{L}_{normal} + \mathcal{L}_{variance} + \mathcal{L}_{rank} + \mathcal{L}_{\gamma} + \mathcal{L}_{proj} + \mathcal{L}_{mask}
\end{equation}
where $\mathcal{L}_{render}$ is the photometric loss, $\mathcal{L}_{normal}$ is the regularization on rendered normals~\cite{huang20242d}, $\mathcal{L}_{variance}$ forces the rendered depth variance to be small~\cite{gao2024relightable}, $\mathcal{L}_{rank}$ forces the Gaussians to be disc-like~\cite{erank}, further improving their geometric fidelity, and $\mathcal{L}_{\gamma}$ and $\mathcal{L}_{proj}$ are the median and projection losses for the SDF~\cite{zhu2025gaussian}.

\subsection{Dynamic Inverse Rendering}
\label{sec:inverse_rendering}
Given a reliable estimate of the geometry from the second stage, we introduce a material model on top of the 3D Gaussian representation and jointly optimize it with the environment illumination to recover accurate, disentangled materials from observations of the object under varying poses.

\noindent\textbf{Material and Environment Representation.}
We parameterize the spatially varying BRDF $f$ with a diffuse and a specular term. The diffuse term is modelled with a Lambertian reflectance model as $f_d = \frac{a}{\pi}$, where $a$ is the albedo, and the specular term is represented by the GGX microfacet model~\cite{walter2007microfacet} as:
\begin{equation}
    f_s(x_t, \hat{\omega}_{it}, \hat{\omega}_o, r, F_0) = \frac{DFG}{4 \cdot (\hat{\omega}_{i}\cdot\hat{n}) \cdot (\hat{\omega}_{o} \cdot \hat{n})}
\end{equation}
where $r$ is the roughness, $F_0$ is the specular reflectance at normal incidence and $D$, $F$ and $G$ are the normal distribution, Fresnel and geometry attenuation functions. The final BRDF is represented as $f = f_d + f_s$. We add the optimizable material parameters $a \in \mathbb{R}^3$ and $r \in \mathbb{R}$ to the Gaussians at this stage, which can be rendered into material maps similar to the other Gaussian properties. We assume dielectric materials with $F_0 = 0.04$. The material is made spatially varying by storing four values for each material property and the normal at the corners of the Gaussian plane, which are then interpolated at the intersection point, as introduced in SVG-IR~\cite{sun2025svg}, thus improving representation capacity and accuracy. We assume that the illumination is distant and can be represented as an environment map. Since a high resolution environment map can be used as an error sink early on in the optimization, we start at a low resolution and bilinearly upsample it progressively to a higher resolution.

\noindent\textbf{Visibility and Global Illumination.}
Since we want to consolidate multiple surface-light interactions into consistent material estimates, it is important to model self-occlusions for an object. We use the ray tracer for Gaussians introduced by 3DGRT~\cite{moenne20243dgrt} to model such occlusions in our pipeline. Given secondary rays originating from surface point $x$ towards sampled incident light directions $\hat{\omega}_i$, the tracer is queried as
\begin{equation}
    (c_i, o_i) \leftarrow Trace(x, \hat{\omega}_i)
\end{equation}
where the visibility $V$ can be deduced by thresholding the rendered opacity $o_i$ and the rendered colour $c_i$ is used as an approximation of the indirect illumination coming from $\omega_i$. Plugging these values into Equation~\ref{eq:illumination} gives us the final incoming illumination.

\noindent\textbf{Rendering and Optimization.}
We adopt the deferred shading approach in our pipeline. We first render the Gaussians, transformed with object pose $(R_t, t_t)$ for timestep $t$, to get the depth, normal and material maps. These maps are then used to perform physically-based rendering at the estimated surface point using Equation~\ref{eq:rendering}.
We use Fibonacci-based sphere sampling for rendering which adds bias of a fixed pattern but provides optimization stability~\cite{yao2022neilf}. The final loss is defined as:
\begin{equation}
    \mathcal{L}_{pbr} = \mathcal{L}_{render} + \mathcal{L}_{light} + \mathcal{L}_{material}
\end{equation}
where $\mathcal{L}_{light}$ and $\mathcal{L}_{material}$ define the smoothness losses on the environment and material maps~\cite{gao2024relightable}.

\noindent\textbf{Relighting.} Since optimization instability is no longer a concern during relighting, we use BRDF- and lighting-based importance sampling~\cite{pharr2023physically} to get high-quality relights. As the optimized radiance of the Gaussians becomes invalid for indirect illumination during relighting, we use the split-sum approximation~\cite{karis2013real} to efficiently compute one-bounce indirect radiance for improved fidelity of relights~\cite{gu2025irgs}.

\section{Experiments}
\label{sec:experiments}

In the following, we describe our experimental setup, extensively evaluate the material-lighting decomposition performance and benchmark our method against existing baselines. See the supplementary material for implementation details.

\noindent\textbf{Synthetic Data.}
HOT3D~\cite{banerjee2025hot3d} provides a commonly used dataset of hand-held object interactions. As evaluating our method requires ground-truth material properties and relit views, we use the object assets from this dataset to generate our own renders with known ground-truth materials and illumination.
Based on these assets, we create the following settings designed to mimic common real-world capture scenarios:
\begin{itemize}
    \item \textbf{Static Dome \includegraphics[scale=0.075]{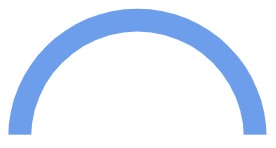}:} The object remains static while the camera moves around it, capturing views from the upper hemisphere.
    
    \item \textbf{Turntable \includegraphics[scale=0.065]{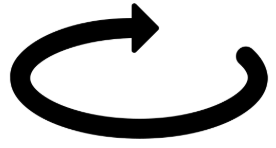}:} This setting models the classical turntable scenario, where the camera is static and the object rotates around a single axis. While this introduces varying illumination conditions, the motion is limited to a single degree of freedom.
    
    \item \textbf{Hand-held Rotations \includegraphics[scale=0.065]{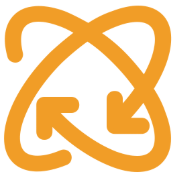}:} 
    This is our proposed setting, where the user holds the object in front of the camera and rotates it around arbitrary axes. In the synthetic data, we do not explicitly model the hand, but instead render the object undergoing arbitrary rotations.
\end{itemize}
We select ten objects with varying geometries and render two material variants: an original variant using the material textures provided with the HOT3D assets, which are generally more specular, and a diffuse variant that keeps the albedo texture but uses a Lambertian material model, where material and lighting are harder to disentangle. The camera trajectories for the three capture settings are designed to share a common set of views, which we use for evaluation.

\noindent\textbf{Real Data.}
For a valid comparison between capture settings, we require the object to be captured under the same lighting conditions for each setting. Since none of the existing real-world benchmarks satisfy this requirement, we record two real hand-held objects in the static dome and hand-held settings under the same environment lighting. We use SAM3~\cite{carion2026sam} to obtain masks for the captured objects.

\noindent\textbf{Metrics.}
We use PSNR, SSIM, and LPIPS for albedo and relighting to assess decomposition quality, mean angular error (MAE) for normals and RMSE for environment maps. We normalize the albedo and relighting scale using ground-truth albedo for synthetic evaluation~\cite{zhang2021nerfactor}.

\subsection{Material-Lighting Decomposition Evaluation}

\begin{figure}[t]
    \centering
    \includegraphics[width=\textwidth]{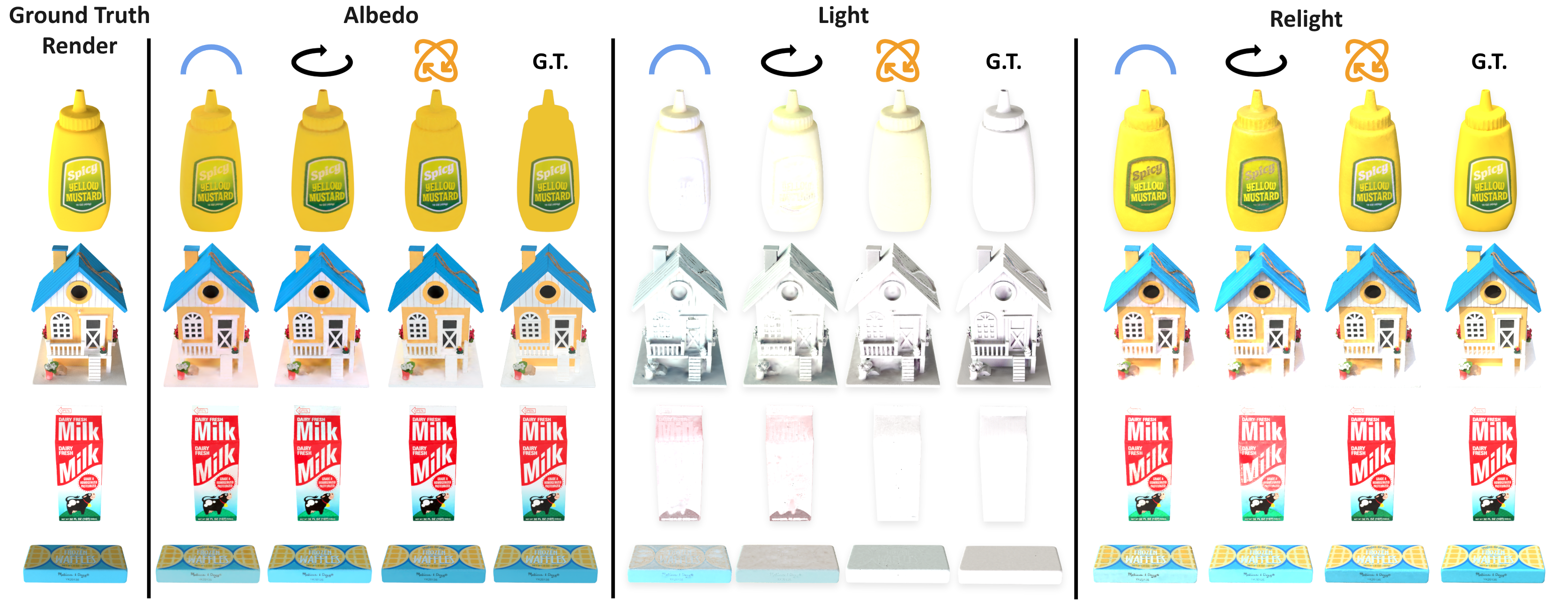}
    \caption{Qualitative results for different capture configurations on our proposed synthetic dataset for the diffuse variant. The improved disentanglement of albedo and lighting (note the progression of disentanglement from left to right) induced by the hand-held setting results in better relighting consistently across a diverse set of objects.}
    \label{fig:synthetic_capture_configurations}
\end{figure}

\begin{table}[t]
    \setlength{\cmidrulewidth}{0.06em}
    \centering
    \caption{Quantitative results for different capture configurations on our proposed dataset. Results are reported for original and diffuse object variants. Despite evaluation with ground-truth poses, the static setting struggles to disentangle material and lighting, especially for the diffuse variant, while the hand-held setting performs well on both variants and with both estimated and ground-truth poses. \textbf{Bold} indicates the best result, and \underline{underline} indicates the second-best result.}
    
    \label{tab:synthetic_capture_configurations}
    \scalebox{1}{
        \begin{minipage}{\textwidth}\centering
            \begin{tabular}{c|c|ccc|ccc|c}
                \toprule
                \multirow{2}{*}{Setting} & \multirow{2}{*}{Pose} & \multicolumn{3}{c|}{Albedo} & \multicolumn{3}{c|}{Relighting} & Normal \\
                & & PSNR$\uparrow$ & SSIM$\uparrow$ & LPIPS$\downarrow$ & PSNR$\uparrow$ & SSIM$\uparrow$ & LPIPS$\downarrow$ & MAE$\downarrow$ \\
                \midrule
                \multicolumn{9}{c}{Original} \\
                \midrule
                \includegraphics[scale=0.05]{figures/symbols_static_dome.png} & G.T. & 36.72 & \underline{0.987} & 0.018 & 38.14 & 0.986 & 0.020 & \underline{0.83} \\ 
                \midrule
                \multirow{2}{*}[-0.75ex]{\includegraphics[scale=0.05]{figures/symbols_turntable.png}} & Est. & 34.10 & 0.983 & 0.021 & 33.02 & 0.977 & 0.027 & 2.45 \\ 
                \cmidrule{2-9}
                & G.T. & 36.15 & \underline{0.987} & \underline{0.016} & 36.01 & 0.983 & 0.022 & 1.20 \\
                \midrule
                \multirow{2}{*}[-0.75ex]{\includegraphics[scale=0.05]{figures/symbols_handheld.png}} & Est. & \underline{40.37} & \textbf{0.991} & \textbf{0.013} & \underline{38.54} & \underline{0.988} & \underline{0.017} & 1.56 \\
                \cmidrule{2-9}
                & G.T. & \textbf{40.72} & \textbf{0.991} & \textbf{0.013} & \textbf{40.24} & \textbf{0.989} & \textbf{0.016} & \textbf{0.53} \\
                \midrule
                \multicolumn{9}{c}{Diffuse} \\
                \midrule
                \includegraphics[scale=0.05]{figures/symbols_static_dome.png} & G.T. & 32.68 & 0.983 & 0.021 & 34.85 & 0.981 & 0.024 & 1.59 \\ 
                \midrule
                \multirow{2}{*}[-0.75ex]{\includegraphics[scale=0.05]{figures/symbols_turntable.png}} & Est. & 32.97 & 0.982 & 0.022 & 33.04 & 0.976 & 0.027 & 1.83 \\ 
                \cmidrule{2-9}
                & G.T. & 35.65 & \underline{0.984} & \underline{0.016} & 35.10 & 0.981 & \underline{0.023} & 1.42 \\
                \midrule
                \multirow{2}{*}[-0.75ex]{\includegraphics[scale=0.05]{figures/symbols_handheld.png}} & Est. & \underline{40.33} & \textbf{0.990} & \textbf{0.010} & \underline{40.37} & \underline{0.990} & \textbf{0.015} & \underline{0.75} \\
                \cmidrule{2-9}
                & G.T. & \textbf{40.55} & \textbf{0.990} & \textbf{0.010} & \textbf{41.29} & \textbf{0.991} & \textbf{0.015} & \textbf{0.58} \\
                \bottomrule
            \end{tabular}
        \end{minipage}
    }
\end{table}

\noindent\textbf{Synthetic Data.}
Figure~\ref{fig:synthetic_capture_configurations} provides a qualitative comparison for the three common capture trajectories reproduced in our synthetic dataset. In the static dome setting, the fixed incoming illumination observed at each surface point leads to a stronger ambiguity, causing the albedo colour to bleed into the illumination. The turntable case increases the surface-light interactions for the surface points but they are still comparatively limited since the rotation is constrained to a single axis. The hand-held rotation setting, in contrast, exposes each surface point to a diverse set of light interactions, which imposes stronger constraints on the proper separation of albedo and lighting, and yields much better renders under novel illumination.

The quantitative results are provided in Table~\ref{tab:synthetic_capture_configurations}. We give full advantage of ground-truth camera poses to the static setting and evaluate the turntable and hand-held settings both with ground-truth and estimated poses. Notably, the hand-held setting with estimated poses outperforms the static setting with ground-truth poses despite having worse normals, highlighting the advantage of using the hand-held setting. In general, the hand-held setting achieves the strongest decomposition performance and the best relighting metrics, supporting our hypothesis that motion helps disentangle material and illumination during inverse rendering. The turntable setting additionally suffers from unconstrained normal reconstruction due to limited view coverage.

\begin{figure}[t]
    \centering
    \includegraphics[width=\textwidth]{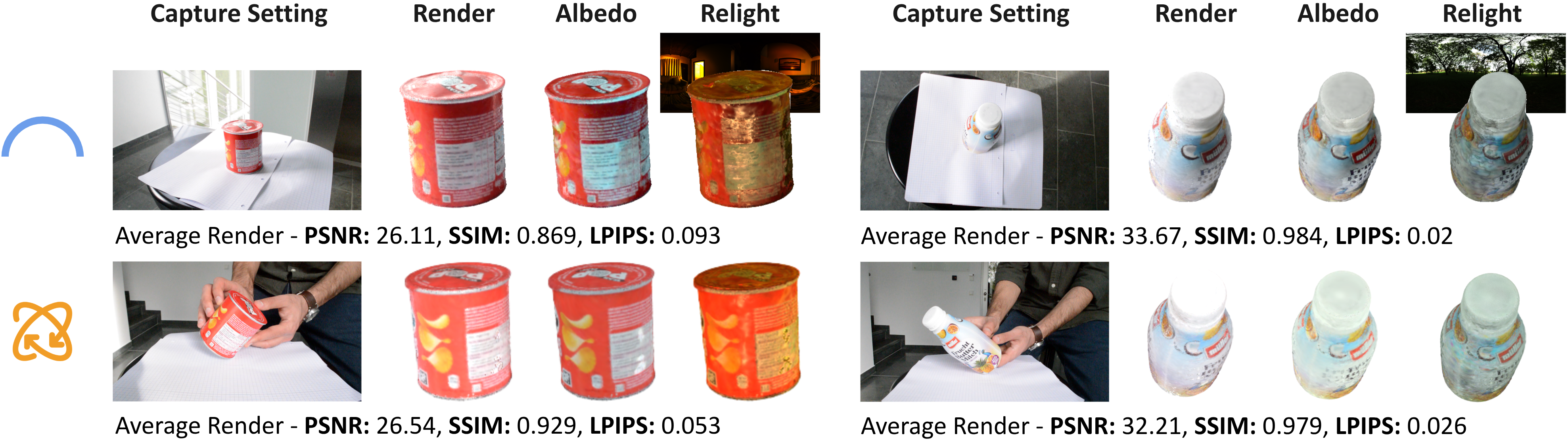}
    \caption{Real captures under static multiview (top) and hand-held (bottom) settings. Despite similar render quality, hand-held capture yields cleaner albedo and more plausible relighting.}
    \label{fig:real}
\end{figure}

\noindent\textbf{Real Data.}
We demonstrate the advantage of the hand-held setting over the common static casual capture setting on two real captured objects in Figure~\ref{fig:real}. Despite similar rendering quality across the full sequences, the decomposition is qualitatively better in the hand-held setting. The static setting bakes shadows from the render into the albedo and tints the white regions, while the hand-held setting produces flatter and more consistent albedo, as expected, which in turn results in more plausible relights. This shows the viability of our method in real-world scenarios.

\begin{figure}[t]
    \centering
    \includegraphics[width=\textwidth]{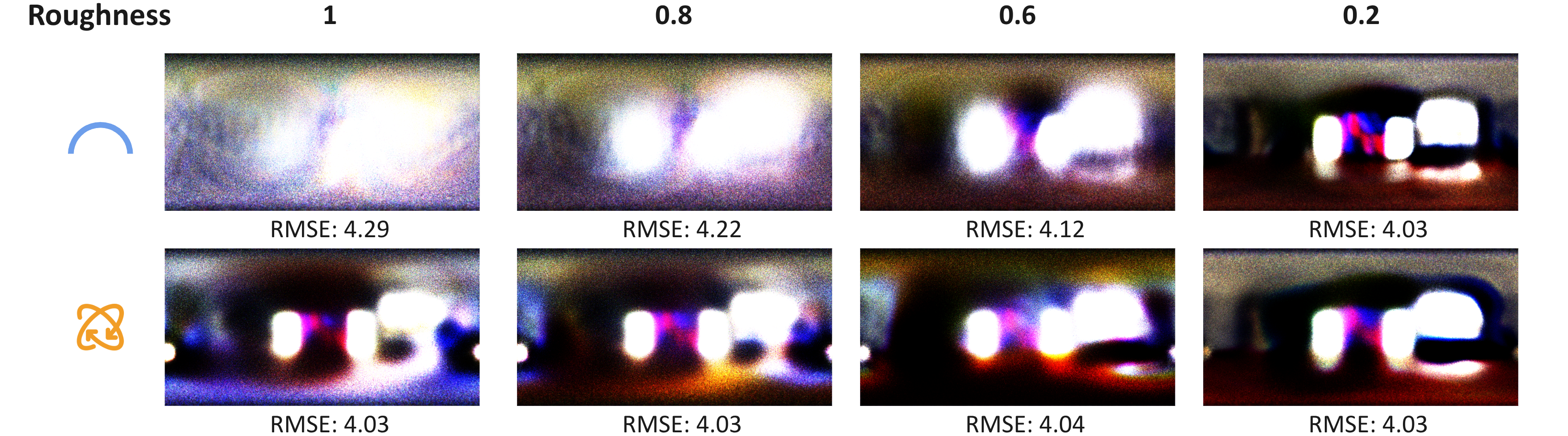}
    \caption{Environment reconstruction under varying roughness. The hand-held setting recovers sharper illumination structure even for diffuse objects, while the gap to the static setting narrows for more specular objects, where sharper highlights provide stronger lighting constraints. The ground truth environment map is shown in Figure~\ref{fig:method}.}
    \label{fig:roughness_analysis}
\end{figure}

\noindent\textbf{Material Roughness Analysis.}
To analyze how the material properties affect decomposition ambiguity, we create different versions of one object in our synthetic dataset by varying the roughness while keeping the albedo fixed. The resulting objects are rendered under the same environment map using the static and hand-held settings.
Figure~\ref{fig:roughness_analysis} shows the reconstructed environment maps from the resulting sequences, which most clearly illustrates the effect of different roughness levels. Lower roughness already helps disambiguation in the static setting since the sharper highlights provide stronger constraints and cannot be explained by the diffuse albedo alone. As the roughness increases and the reflections become more diffuse, it becomes harder to recover the illumination signal, increasing the tendency of the static setting to trade off albedo and illumination. The diverse surface-light interactions in the hand-held setting provide stronger directionality---which also curbs the swapping tendency of albedo and illumination---and enable the model to recover more information even in the challenging diffuse case while retaining the benefit across the roughness range.

\begin{figure}[t]
    \centering
    \includegraphics[width=\textwidth]{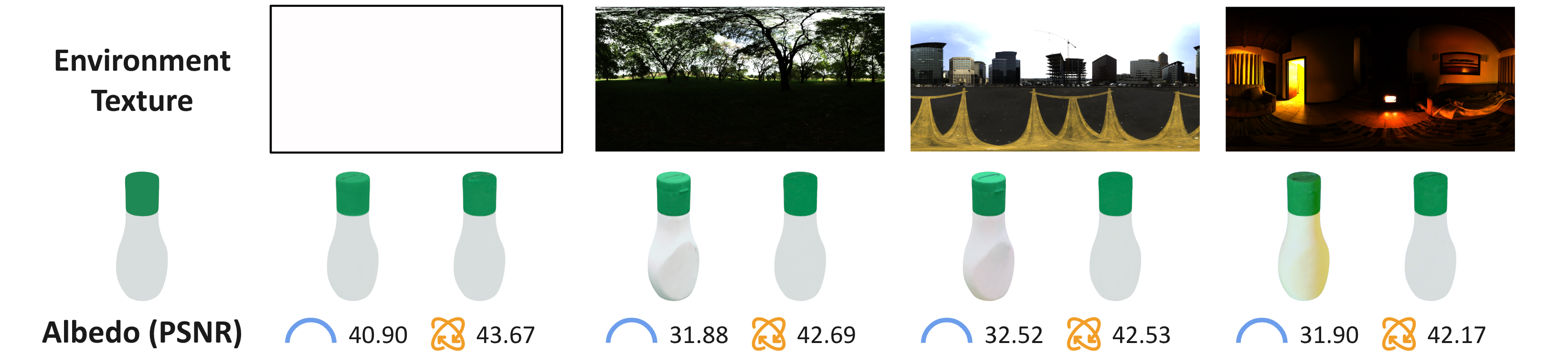}
    \caption{Effect of illumination complexity on decomposition. The hand-held setting maintains high albedo quality under increasingly complex lighting, while the static setting degrades.}
    \label{fig:illumination_analysis}
\end{figure}

\noindent\textbf{Illumination Analysis.}
We further study the effect of different environment maps on the material decomposition quality for the static and hand-held settings. Figure~\ref{fig:illumination_analysis} illustrates this by reconstructing the object from views rendered with different environment maps. In the case of uniform white light, no additional surface-light interactions are available to the hand-held setting, while the estimated albedo from the static setting degrades only mildly due to the reduced lighting complexity. As we progress to environments with more varying lighting, the albedo disentanglement starts degrading in the static setting due to increased lighting complexity while the hand-held setting, due to the stronger constraints from increasingly diverse surface-light interactions, is able to maintain the albedo quality.

\subsection{Benchmarking Relighting Performance}

In our work, we introduce the new hand-held setting that requires object tracking. We validate our design choices in two ways: (1) performance in the static setting on a standard benchmark and (2) performance in the dynamic setting against adapted baselines. See the supplementary material for our pose estimation performance.

\noindent\textbf{Static Setting.}
Table~\ref{tab:benchmark_static} shows the evaluation of albedo reconstruction, relighting under novel illumination and normal accuracy on the standard TensoIR benchmark~\cite{jin2023tensoir}. Our method is competitive with baselines across metrics, achieving improved albedo and relighting quality. This suggests that our pipeline is already capable of achieving the strongest decomposition results for static inverse rendering, and motivates using the same pipeline in the dynamic setting where additional motion-induced lighting constraints further help improve the decomposition.

\noindent\textbf{Dynamic Setting.}
As there are no existing baselines for test-time dynamic inverse rendering of general objects, we adapt existing methods for comparison on our dataset in Table~\ref{tab:benchmark_dynamic}.
ReCap~\cite{li2025recap} optimizes a separate environment map for each capture illumination. We treat each frame of a dynamic scene as a separate lighting condition and optimize a separate environment for it (shared over consecutive frames for efficiency). Such a setup is not suitable for dynamic scenes: sharing more frames per environment approximates too much while less sharing runs into optimization issues. Designed for static scenes, we extend SVG-IR~\cite{sun2025svg} and IRGS~\cite{gu2025irgs} to account for object dynamics when querying the environment map. Our consolidated SDF-based geometry modelling yields the strongest decomposition and subsequent relighting performance in the hand-held setting.

\begin{table}[t]
    \centering

    \begin{minipage}[t]{0.42\textwidth}
        \centering
        \caption{Decomposition results on the synthetic TensoIR dataset, where we outperform the static-setting baselines.}
        \label{tab:benchmark_static}
        \vspace{3.5mm}
        \scalebox{1}{%
            \begin{tabular}{c|c|c|c}
                \toprule
                \multirow{2}{*}{Method} & Albedo & Relighting & Normal \\
                & PSNR $\uparrow$ & PSNR $\uparrow$ & MAE $\downarrow$ \\
                \midrule
                SVG-IR & 30.27 & 31.19 & 4.19 \\
                \midrule
                IRGS & 33.41 & 30.63 & \textbf{3.99} \\
                \midrule
                Ours & \textbf{33.96} & \textbf{32.17} & 4.19 \\
                \bottomrule
            \end{tabular}
        }
    \end{minipage}
    \hfill
    \begin{minipage}[t]{0.55\textwidth}
        \centering
        \caption{Decomposition results on four hand-held sequences of our dataset. *Extended to handle rotating objects.}
        \label{tab:benchmark_dynamic}
        \scalebox{1}{%
            \begin{tabular}{c|c|c|c|c}
                \toprule
                \multirow{2}{*}{Method} & \multirow{2}{*}{Envs} & Albedo & Relighting & Normal \\
                & & PSNR$\uparrow$ & PSNR$\uparrow$ & MAE$\downarrow$ \\
                \midrule
                \multirow{3}{*}{ReCap} & 10 & 21.91 & 25.74 & 8.06 \\
                & 40 & 22.42 & 25.58 & 8.01 \\
                & 100 & 22.31 & 25.20 & 8.08 \\
                \midrule
                SVG-IR* & 1 & 33.40 & 32.22 & 3.69 \\ 
                \midrule
                IRGS* & 1 & 40.26 & 32.99 & 0.56 \\
                \midrule
                Ours & 1 & \textbf{41.03} & \textbf{39.65} & \textbf{0.51} \\
                \bottomrule
            \end{tabular}
        }
    \end{minipage}

\end{table}

\section{Conclusion}
\label{sec:conclusion}

Instead of introducing additional light sources or relying on strong learned priors, we investigate the long-standing material-lighting ambiguity in inverse rendering from the perspective of leveraging the natural diversity of surface-light interactions induced by rigid object motion. Our proposed relightable 4D reconstruction pipeline leverages these interactions to improve decomposition in the challenging hand-held capture setting. Across controlled experiments on our proposed synthetic dataset, we show that increasing object pose diversity yields stronger constraints for disentangling albedo and illumination, while we also demonstrate the feasibility of our method in improving decomposition for real hand-held objects.

\noindent\textbf{Future Work.} Modelling hand occlusions and more complex near-field light interactions in real scenes is an exciting future direction. Moreover, non-rigid objects can provide even richer motion diversity, making a general non-rigid inverse rendering pipeline a natural next step.


\section*{Acknowledgements}
Jan Eric Lenssen is supported by the German Research Foundation (DFG) - 556415750 (Emmy Noether Programme, project: Spatial Modeling and Reasoning).

%
%
\bibliographystyle{splncs04}
\bibliography{main}

\begin{thebibliography}{10}
\providecommand{\url}[1]{\texttt{#1}}
\providecommand{\urlprefix}{URL }
\providecommand{\doi}[1]{https://doi.org/#1}

\bibitem{alzayer2025genrelight}
Alzayer, H., Henzler, P., Barron, J.T., Huang, J.B., Srinivasan, P.P., Verbin, D.: Generative multiview relighting for 3d reconstruction under extreme illumination variation. In: IEEE/CVF Conference on Computer Vision and Pattern Recognition (CVPR). pp. 10933--10942 (June 2025)

\bibitem{banerjee2025hot3d}
Banerjee, P., Shkodrani, S., Moulon, P., Hampali, S., Han, S., Zhang, F., Zhang, L., Fountain, J., Miller, E., Basol, S., Newcombe, R., Wang, R., Engel, J.J., Hodan, T.: Hot3d: Hand and object tracking in 3d from egocentric multi-view videos. In: IEEE/CVF Conference on Computer Vision and Pattern Recognition (CVPR). pp. 7061--7071 (June 2025)

\bibitem{boss2021nerd}
Boss, M., Braun, R., Jampani, V., Barron, J.T., Liu, C., Lensch, H.: Nerd: Neural reflectance decomposition from image collections. In: IEEE/CVF International Conference on Computer Vision (ICCV). pp. 12684--12694 (2021)

\bibitem{boss2022samurai}
Boss, M., Engelhardt, A., Kar, A., Li, Y., Sun, D., Barron, J., Lensch, H., Jampani, V.: Samurai: Shape and material from unconstrained real-world arbitrary image collections. Advances in Neural Information Processing Systems (NeurIPS)  \textbf{35},  26389--26403 (2022)

\bibitem{carion2026sam}
Carion, N., Gustafson, L., Hu, Y.T., Debnath, S., Hu, R., Coll-Vinent, D.S., Ryali, C., Alwala, K.V., Khedr, H., Huang, A., Lei, J., Ma, T., Guo, B., Kalla, A., Marks, M., Greer, J., Wang, M., Sun, P., R{\"a}dle, R., Afouras, T., Mavroudi, E., Xu, K., Wu, T.H., Zhou, Y., Momeni, L., HAZRA, R., Ding, S., Vaze, S., Porcher, F., Li, F., Li, S., Kamath, A., Cheng, H.K., Dollar, P., Ravi, N., Saenko, K., Zhang, P., Feichtenhofer, C.: {SAM} 3: Segment anything with concepts. In: International Conference on Learning Representations (ICLR) (2026), \url{https://openreview.net/forum?id=r35clVtGzw}

\bibitem{chen2022tensorf}
Chen, A., Xu, Z., Geiger, A., Yu, J., Su, H.: Tensorf: Tensorial radiance fields. In: European Conference on Computer Vision (ECCV) (2022)

\bibitem{chen2024intrinsicanything}
Chen, X., Peng, S., Yang, D., Liu, Y., Pan, B., Lv, C., Zhou, X.: Intrinsicanything: Learning diffusion priors for inverse rendering under unknown illumination. In: European Conference on Computer Vision (ECCV). p. 450–467. Springer-Verlag, Berlin, Heidelberg (2024). \doi{10.1007/978-3-031-73027-6_26}, \url{https://doi.org/10.1007/978-3-031-73027-6_26}

\bibitem{chen2025hort}
Chen, Z., Potamias, R.A., Chen, S., Schmid, C.: {HORT}: Monocular hand-held objects reconstruction with transformers. In: IEEE/CVF International Conference on Computer Vision (ICCV) (2025)

\bibitem{dihlmann2024subsurface}
Dihlmann, J.N., Majumdar, A., Engelhardt, A., Braun, R., Lensch, H.: Subsurface scattering for gaussian splatting. Advances in Neural Information Processing Systems (NeurIPS)  \textbf{37},  121765--121789 (2024)

\bibitem{edstedt2025romav2}
Edstedt, J., Nordström, D., Zhang, Y., Bökman, G., Astermark, J., Larsson, V., Heyden, A., Kahl, F., Wadenbäck, M., Felsberg, M.: Roma v2: Harder better faster denser feature matching (2025), \url{https://arxiv.org/abs/2511.15706}

\bibitem{engelhardt2024shinobi}
Engelhardt, A., Raj, A., Boss, M., Zhang, Y., Kar, A., Li, Y., Sun, D., Brualla, R.M., Barron, J.T., Lensch, H., et~al.: Shinobi: Shape and illumination using neural object decomposition via brdf optimization in-the-wild. In: IEEE/CVF Conference on Computer Vision and Pattern Recognition (CVPR). pp. 19636--19646 (2024)

\bibitem{fan2024hold}
Fan, Z., Parelli, M., Kadoglou, M.E., Chen, X., Kocabas, M., Black, M.J., Hilliges, O.: Hold: Category-agnostic 3d reconstruction of interacting hands and objects from video. In: IEEE/CVF Conference on Computer Vision and Pattern Recognition (CVPR). pp. 494--504 (2024)

\bibitem{fei2024vminer}
Fei, F., Tang, J., Tan, P., Shi, B.: Vminer: Versatile multi-view inverse rendering with near- and far-field light sources. In: IEEE/CVF Conference on Computer Vision and Pattern Recognition (CVPR). pp. 11800--11809 (June 2024)

\bibitem{fischer2025unified}
Fischer, T., Zhang, X., Ilg, E.: Unified category-level object detection and pose estimation from rgb images using 3d prototypes. In: IEEE/CVF International Conference on Computer Vision (ICCV). pp. 9790--9800 (2025)

\bibitem{colmap-free-gs}
Fu, Y., Wang, X., Liu, S., Kulkarni, A., Kautz, J., Efros, A.A.: Colmap-free 3d gaussian splatting. In: {CVPR}. pp. 20796--20805. {IEEE} (2024)

\bibitem{gao2024relightable}
Gao, J., Gu, C., Lin, Y., Li, Z., Zhu, H., Cao, X., Zhang, L., Yao, Y.: Relightable 3d gaussians: Realistic point cloud relighting with brdf decomposition and ray tracing. In: Leonardis, A., Ricci, E., Roth, S., Russakovsky, O., Sattler, T., Varol, G. (eds.) European Conference on Computer Vision (ECCV). pp. 73--89. Springer Nature Switzerland, Cham (2025)

\bibitem{gu2025irgs}
Gu, C., Wei, X., Zeng, Z., Yao, Y., Zhang, L.: Irgs: Inter-reflective gaussian splatting with 2d gaussian ray tracing. In: IEEE/CVF Conference on Computer Vision and Pattern Recognition (CVPR). pp. 10943--10952 (2025)

\bibitem{gu2025tensoflow}
Gu, C., Wei, X., Zhang, L., Zhu, X.: Tensoflow: Tensorial flow-based sampler for inverse rendering. In: IEEE/CVF Conference on Computer Vision and Pattern Recognition (CVPR). pp. 495--504 (2025)

\bibitem{hampali2020honnotate}
Hampali, S., Rad, M., Oberweger, M., Lepetit, V.: Honnotate: A method for 3d annotation of hand and object poses. In: IEEE/CVF Conference on Computer Vision and Pattern Recognition (CVPR) (2020)

\bibitem{hasselgren2022nvdiffrecmc}
Hasselgren, J., Hofmann, N., Munkberg, J.: Shape, light, and material decomposition from images using monte carlo rendering and denoising. In: Koyejo, S., Mohamed, S., Agarwal, A., Belgrave, D., Cho, K., Oh, A. (eds.) Advances in Neural Information Processing Systems (NeurIPS). vol.~35, pp. 22856--22869. Curran Associates, Inc. (2022), \url{https://proceedings.neurips.cc/paper_files/paper/2022/file/8fcb27984bf16ca03cad643244ec470d-Paper-Conference.pdf}

\bibitem{huang20242d}
Huang, B., Yu, Z., Chen, A., Geiger, A., Gao, S.: 2d gaussian splatting for geometrically accurate radiance fields. In: ACM SIGGRAPH. SIGGRAPH '24, Association for Computing Machinery, New York, NY, USA (2024). \doi{10.1145/3641519.3657428}, \url{https://doi.org/10.1145/3641519.3657428}

\bibitem{huang2025onlinesplatter}
Huang, M.H., Foo, L.G., Theobalt, C., Sun, Y., Soh, D.W.: Onlinesplatter: Pose-free online 3d reconstruction for free-moving objects. In: Advances in Neural Information Processing Systems (NeurIPS) (2025)

\bibitem{erank}
Hyung, J., Hong, S., Hwang, S., Lee, J., Choo, J., Kim, J.: Effective rank analysis and regularization for enhanced 3d gaussian splatting. In: NeurIPS (2024)

\bibitem{jiang2024gaussianshader}
Jiang, Y., Tu, J., Liu, Y., Gao, X., Long, X., Wang, W., Ma, Y.: Gaussianshader: 3d gaussian splatting with shading functions for reflective surfaces. In: IEEE/CVF Conference on Computer Vision and Pattern Recognition (CVPR). pp. 5322--5332 (2024)

\bibitem{jin2023tensoir}
Jin, H., Liu, I., Xu, P., Zhang, X., Han, S., Bi, S., Zhou, X., Xu, Z., Su, H.: Tensoir: Tensorial inverse rendering. In: Proceedings of the IEEE/CVF Conference on Computer Vision and Pattern Recognition (CVPR). pp. 165--174 (June 2023)

\bibitem{jin20256dope}
Jin, Y., Prasad, V., Jauhri, S., Franzius, M., Chalvatzaki, G.: 6dope-gs: Online 6d object pose estimation using gaussian splatting. In: IEEE/CVF International Conference on Computer Vision (ICCV). pp. 8032--8043 (2025)

\bibitem{kajiya1986rendering}
Kajiya, J.T.: The rendering equation. ACM SIGGRAPH Computer Graphics  \textbf{20}(4),  143--150 (1986). \doi{10.1145/15886.15902}

\bibitem{karis2013real}
Karis, B., Games, E.: Real shading in unreal engine 4. Proc. Physically Based Shading Theory Practice  \textbf{4}(3), ~1 (2013)

\bibitem{kerbl20233d}
Kerbl, B., Kopanas, G., Leimk{\"u}hler, T., Drettakis, G.: 3d gaussian splatting for real-time radiance field rendering. ACM Transactions on Graphics (ToG)  \textbf{42}(4),  1--14 (2023)

\bibitem{kuang2022neroic}
Kuang, Z., Olszewski, K., Chai, M., Huang, Z., Achlioptas, P., Tulyakov, S.: Neroic: Neural rendering of objects from online image collections. ACM Transactions on Graphics (ToG)  \textbf{41}(4),  1--12 (2022)

\bibitem{leroy2024grounding}
Leroy, V., Cabon, Y., Revaud, J.: Grounding image matching in 3d with mast3r. In: European Conference on Computer Vision (ECCV). p. 71–91. Springer-Verlag, Berlin, Heidelberg (2024). \doi{10.1007/978-3-031-73220-1_5}, \url{https://doi.org/10.1007/978-3-031-73220-1_5}

\bibitem{li2024tensosdf}
Li, J., Wang, L., Zhang, L., Wang, B.: Tensosdf: Roughness-aware tensorial representation for robust geometry and material reconstruction. ACM Transactions on Graphics (ToG)  \textbf{43}(4) (Jul 2024). \doi{10.1145/3658211}, \url{https://doi.org/10.1145/3658211}

\bibitem{li2025recap}
Li, J., Wu, Z., Zamfir, E., Timofte, R.: Recap: Better gaussian relighting with cross-environment captures. In: IEEE/CVF Conference on Computer Vision and Pattern Recognition (CVPR). pp. 21307--21316 (June 2025)

\bibitem{liang2024gs}
Liang, Z., Zhang, Q., Feng, Y., Shan, Y., Jia, K.: Gs-ir: 3d gaussian splatting for inverse rendering. In: IEEE/CVF Conference on Computer Vision and Pattern Recognition (CVPR). pp. 21644--21653 (2024)

\bibitem{liu2023nero}
Liu, Y., Wang, P., Lin, C., Long, X., Wang, J., Liu, L., Komura, T., Wang, W.: Nero: Neural geometry and brdf reconstruction of reflective objects from multiview images. ACM Transactions on Graphics (ToG)  \textbf{42}(4),  1--22 (2023)

\bibitem{lyu2023diffusion}
Lyu, L., Tewari, A., Habermann, M., Saito, S., Zollh\"{o}fer, M., Leimk\"{u}hler, T., Theobalt, C.: Diffusion posterior illumination for ambiguity-aware inverse rendering. ACM Transactions on Graphics (ToG)  \textbf{42}(6) (Dec 2023). \doi{10.1145/3618357}, \url{https://doi.org/10.1145/3618357}

\bibitem{mildenhall2020nerf}
Mildenhall, B., Srinivasan, P.P., Tancik, M., Barron, J.T., Ramamoorthi, R., Ng, R.: Nerf: Representing scenes as neural radiance fields for view synthesis. In: European Conference on Computer Vision (ECCV) (2020)

\bibitem{moenne20243dgrt}
Moenne-Loccoz, N., Mirzaei, A., Perel, O., de~Lutio, R., Martinez~Esturo, J., State, G., Fidler, S., Sharp, N., Gojcic, Z.: 3d gaussian ray tracing: Fast tracing of particle scenes. ACM Transactions on Graphics (ToG)  \textbf{43}(6) (Nov 2024). \doi{10.1145/3687934}, \url{https://doi.org/10.1145/3687934}

\bibitem{mueller2022instant}
M\"{u}ller, T., Evans, A., Schied, C., Keller, A.: Instant neural graphics primitives with a multiresolution hash encoding. ACM Transactions on Graphics (ToG)  \textbf{41}(4) (Jul 2022). \doi{10.1145/3528223.3530127}, \url{https://doi.org/10.1145/3528223.3530127}

\bibitem{pharr2023physically}
Pharr, M., Jakob, W., Humphreys, G.: Physically based rendering: From theory to implementation. MIT Press (2023)

\bibitem{poirier2024diffusion}
Poirier-Ginter, Y., Gauthier, A., Phillip, J., Lalonde, J.F., Drettakis, G.: A diffusion approach to radiance field relighting using multi-illumination synthesis. Computer Graphics Forum  \textbf{43}(4),  e15147 (2024). \doi{https://doi.org/10.1111/cgf.15147}, \url{https://onlinelibrary.wiley.com/doi/abs/10.1111/cgf.15147}

\bibitem{sylvia2006material}
Pont, S.C., te~Pas, S.F.: Material — illumination ambiguities and the perception of solid objects. Perception  \textbf{35}(10),  1331--1350 (2006). \doi{10.1068/p5440}, \url{https://doi.org/10.1068/p5440}, pMID: 17214380

\bibitem{ren2026fastgs}
Ren, S., Wen, T., Fang, Y., Lu, B.: Fastgs: Training 3d gaussian splatting in 100 seconds. In: Proceedings of the IEEE/CVF Conference on Computer Vision and Pattern Recognition. pp. 26094--26103 (2026)

\bibitem{shi2025fmov}
Shi, H., Hu, Y., Koguciuk, D., Lin, J.T., Salzmann, M., Ferstl, D.: Free-moving object reconstruction and pose estimation with virtual camera. AAAI Conference on Artificial Intelligence  \textbf{39}(7),  6860--6868 (Apr 2025). \doi{10.1609/aaai.v39i7.32736}, \url{https://ojs.aaai.org/index.php/AAAI/article/view/32736}

\bibitem{sun2025svg}
Sun, H., Gao, Y., Xie, J., Yang, J., Wang, B.: Svg-ir: Spatially-varying gaussian splatting for inverse rendering. In: IEEE/CVF Conference on Computer Vision and Pattern Recognition (CVPR). pp. 16143--16152 (2025)

\bibitem{sun2021loftr}
Sun, J., Shen, Z., Wang, Y., Bao, H., Zhou, X.: Loftr: Detector-free local feature matching with transformers. In: IEEE/CVF Conference on Computer Vision and Pattern Recognition (CVPR). pp. 8922--8931 (2021)

\bibitem{tang2025rogr}
Tang, J., Levine, M., Verbin, D., Garbin, S.J., Niessner, M., Martin-Brualla, R., Srinivasan, P.P., Henzler, P.: {ROGR: Relightable 3D Objects using Generative Relighting}. In: Advances in Neural Information Processing Systems (NeurIPS) (2025)

\bibitem{tian2022large}
Tian, X., Lin, X., Zhong, F., Qin, X.: Large-displacement 3d object tracking with hybrid non-local optimization. In: European Conference on Computer Vision (ECCV). pp. 627--643. Springer (2022)

\bibitem{verbin2024eclipse}
Verbin, D., Mildenhall, B., Hedman, P., Barron, J.T., Zickler, T., Srinivasan, P.P.: Eclipse: Disambiguating illumination and materials using unintended shadows. In: IEEE/CVF Conference on Computer Vision and Pattern Recognition (CVPR). pp. 77--86 (June 2024)

\bibitem{walter2007microfacet}
Walter, B., Marschner, S.R., Li, H., Torrance, K.E.: Microfacet models for refraction through rough surfaces. Rendering techniques  \textbf{2007}, ~18th (2007)

\bibitem{wang20206}
Wang, C., Mart{\'\i}n-Mart{\'\i}n, R., Xu, D., Lv, J., Lu, C., Fei-Fei, L., Savarese, S., Zhu, Y.: 6-pack: Category-level 6d pose tracker with anchor-based keypoints. In: IEEE International Conference on Robotics and Automation (ICRA). pp. 10059--10066. IEEE (2020)

\bibitem{wang2023deep}
Wang, L., Yan, S., Zhen, J., Liu, Y., Zhang, M., Zhang, G., Zhou, X.: Deep active contours for real-time 6-dof object tracking. In: IEEE/CVF International Conference on Computer Vision (ICCV). pp. 14034--14044 (2023)

\bibitem{wang2021neus}
Wang, P., Liu, L., Liu, Y., Theobalt, C., Komura, T., Wang, W.: Neus: Learning neural implicit surfaces by volume rendering for multi-view reconstruction. In: Ranzato, M., Beygelzimer, A., Dauphin, Y.N., Liang, P., Vaughan, J.W. (eds.) Advances in Neural Information Processing Systems (NeurIPS). pp. 27171--27183 (2021)

\bibitem{wen2020se}
Wen, B., Mitash, C., Ren, B., Bekris, K.E.: se (3)-tracknet: Data-driven 6d pose tracking by calibrating image residuals in synthetic domains. In: IEEE/RSJ International Conference on Intelligent Robots and Systems (IROS). pp. 10367--10373. IEEE (2020)

\bibitem{wen2023bundlesdf}
Wen, B., Tremblay, J., Blukis, V., Tyree, S., M{\"u}ller, T., Evans, A., Fox, D., Kautz, J., Birchfield, S.: Bundlesdf: Neural 6-dof tracking and 3d reconstruction of unknown objects. In: IEEE/CVF Conference on Computer Vision and Pattern Recognition (CVPR). pp. 606--617 (2023)

\bibitem{wen2024foundationpose}
Wen, B., Yang, W., Kautz, J., Birchfield, S.: Foundationpose: Unified 6d pose estimation and tracking of novel objects. In: IEEE/CVF Conference on Computer Vision and Pattern Recognition (CVPR). pp. 17868--17879 (2024)

\bibitem{wu2023nefii}
Wu, H., Hu, Z., Li, L., Zhang, Y., Fan, C., Yu, X.: Nefii: Inverse rendering for reflectance decomposition with near-field indirect illumination. In: IEEE/CVF Conference on Computer Vision and Pattern Recognition (CVPR). pp. 4295--4304 (2023)

\bibitem{yao2022neilf}
Yao, Y., Zhang, J., Liu, J., Qu, Y., Fang, T., McKinnon, D., Tsin, Y., Quan, L.: Neilf: Neural incident light field for physically-based material estimation. In: European Conference on Computer Vision (ECCV). pp. 700--716. Springer (2022)

\bibitem{yariv2021volsdf}
Yariv, L., Gu, J., Kasten, Y., Lipman, Y.: Volume rendering of neural implicit surfaces. In: Ranzato, M., Beygelzimer, A., Dauphin, Y., Liang, P., Vaughan, J.W. (eds.) Advances in Neural Information Processing Systems (NeurIPS). vol.~34, pp. 4805--4815. Curran Associates, Inc. (2021), \url{https://proceedings.neurips.cc/paper_files/paper/2021/file/25e2a30f44898b9f3e978b1786dcd85c-Paper.pdf}

\bibitem{yu2024gsdf}
Yu, M., Lu, T., Xu, L., Jiang, L., Xiangli, Y., Dai, B.: Gsdf: 3dgs meets sdf for improved neural rendering and reconstruction. Advances in Neural Information Processing Systems (NeurIPS)  \textbf{37},  129507--129530 (2024)

\bibitem{radegs}
Zhang, B., Fang, C., Shrestha, R., Liang, Y., Long, X., Tan, P.: Rade-gs: Rasterizing depth in gaussian splatting. CoRR  \textbf{abs/2406.01467} (2024)

\bibitem{zhang2023neilf++}
Zhang, J., Yao, Y., Li, S., Liu, J., Fang, T., McKinnon, D., Tsin, Y., Quan, L.: Neilf++: Inter-reflectable light fields for geometry and material estimation. In: IEEE/CVF International Conference on Computer Vision (ICCV). pp. 3601--3610 (2023)

\bibitem{zhang2021nerfactor}
Zhang, X., Srinivasan, P.P., Deng, B., Debevec, P., Freeman, W.T., Barron, J.T.: Nerfactor: Neural factorization of shape and reflectance under an unknown illumination. ACM Transactions on Graphics (ToG)  \textbf{40}(6),  1--18 (2021)

\bibitem{zhao2024illuminerf}
Zhao, X., Srinivasan, P.P., Verbin, D., Park, K., Martin-Brualla, R., Henzler, P.: Illuminerf: 3d relighting without inverse rendering. In: Globerson, A., Mackey, L., Belgrave, D., Fan, A., Paquet, U., Tomczak, J., Zhang, C. (eds.) Advances in Neural Information Processing Systems (NeurIPS). vol.~37, pp. 42593--42617. Curran Associates, Inc. (2024). \doi{10.52202/079017-1349}

\bibitem{zhu2025gaussian}
Zhu, Z.L., Yang, J., Wang, B.: Gaussian splatting with discretized sdf for relightable assets. In: IEEE/CVF International Conference on Computer Vision (ICCV). pp. 25155--25164 (2025)

\end{thebibliography}

\appendix

\section{Implementation Details}

\noindent\textbf{Stage 1.} We mostly follow the settings from FMOV~\cite{shi2025fmov} for this stage. The geometry is initially warmed up for 100 iterations from the first frame and for each subsequent frame, the pose is first optimized for 500 iterations followed by 500 iterations of joint pose and geometry optimization until all frames are processed. Since we learn coarse geometry at this stage, only 32 samples are taken along each ray for rendering. The learning rate for the pose MLPs is set at 0.0005. The weights corresponding to $\{\mathcal{L}_{eikonal}, \mathcal{L}_{mask}, \mathcal{L}_{match}\}$ are $\{0.1, 5, 0.1\}$.

\noindent\textbf{Stage 2.} At this stage, the global pose MLP is initialized from the learned poses from the previous stage and the Gaussians are initialized on a unit sphere~\cite{zhu2025gaussian}. The poses and geometry are jointly optimized for 150,000 iterations. For synthetic evaluation with ground-truth poses, we skip the first stage and directly plug in the poses at this stage. The default learning rates from 3D-GS~\cite{kerbl20233d} are used. The weights corresponding to $\{\mathcal{L}_{normal}, \mathcal{L}_{variance}, \mathcal{L}_{rank},$ $\mathcal{L}_{\gamma}, \mathcal{L}_{proj}, \mathcal{L}_{mask}\}$ are $\{0.05, 1, 0.1, 5, 1, 1\}$.

\noindent\textbf{Stage 3.} We introduce material parameters at this stage and optimize them jointly with the environment, geometry and poses (after a warm-up of 5000 iterations) for a further 150,000 iterations to let the model fully converge. The geometry is optimized with the learning rate settings from SVG-IR~\cite{sun2025svg}. The learning rate for the material properties is set at 0.025, and the learning rate for the environment map is set at 0.01. The albedo and roughness use a sigmoid activation with a temperature of 2 to allow more gradient flow towards the extremes. The starting resolution of the environment map is $16 \times 32$ and it is upsampled at a regular interval until it reaches the resolution of $1024 \times 2048$. We take 512 Fibonacci sphere samples during optimization while we take 512 samples each from both the BRDF and light distributions during relighting. The weights corresponding to $\{\mathcal{L}_{light}, \mathcal{L}_{material}\}$ are $\{0.05, 0.25\}$.

\begin{figure}[t]
    \centering
    \includegraphics[width=0.8\textwidth]{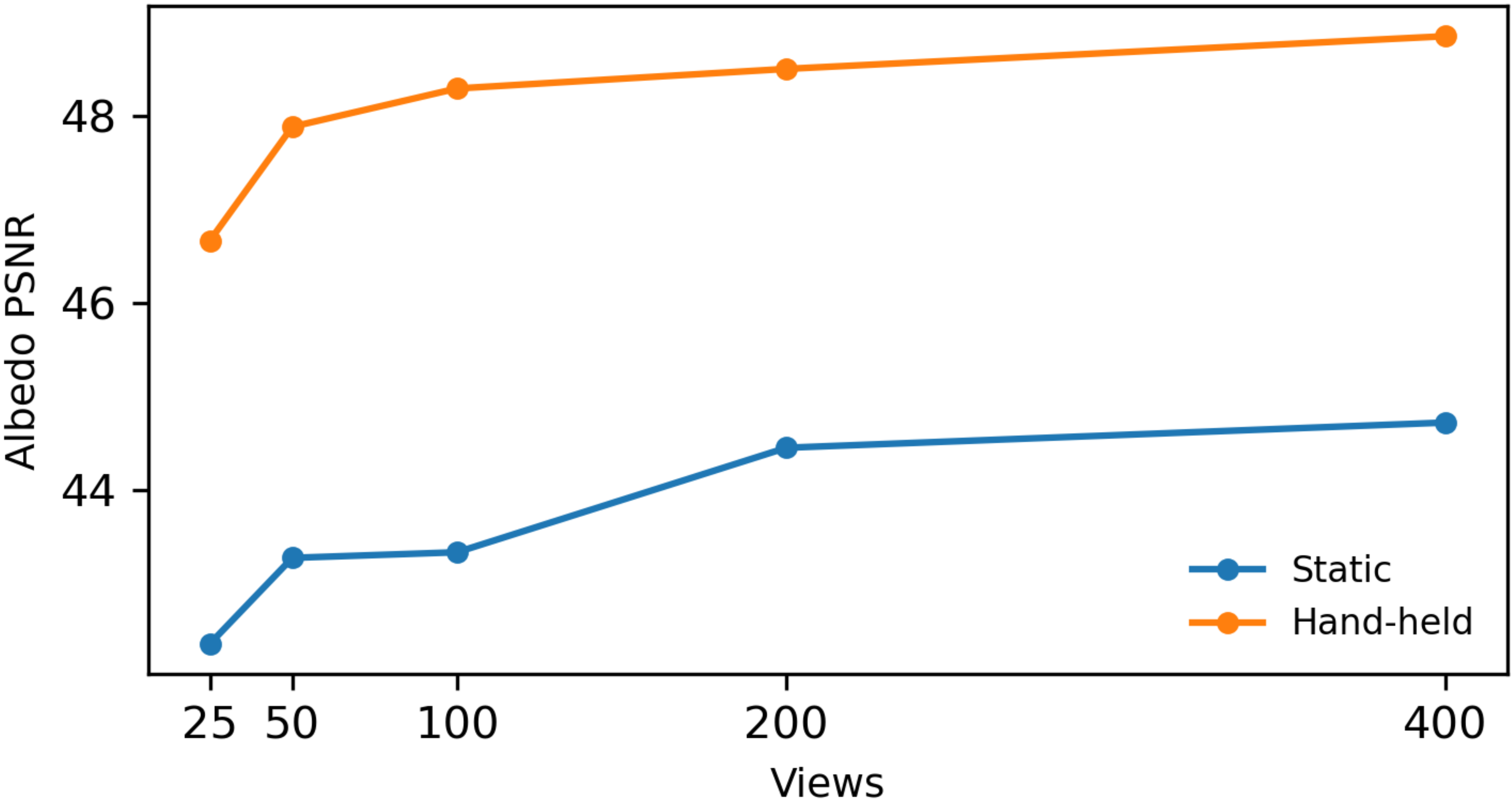}
    \caption{Effect of view sampling on decomposition. A small number of hand-held views outperforms many more static views, showing that the gain comes from motion-induced surface--light interactions rather than view count alone.}
    \label{fig:view_ablation}
\end{figure}

\begin{figure}[t!]
    \centering
    \includegraphics[width=0.8\textwidth]{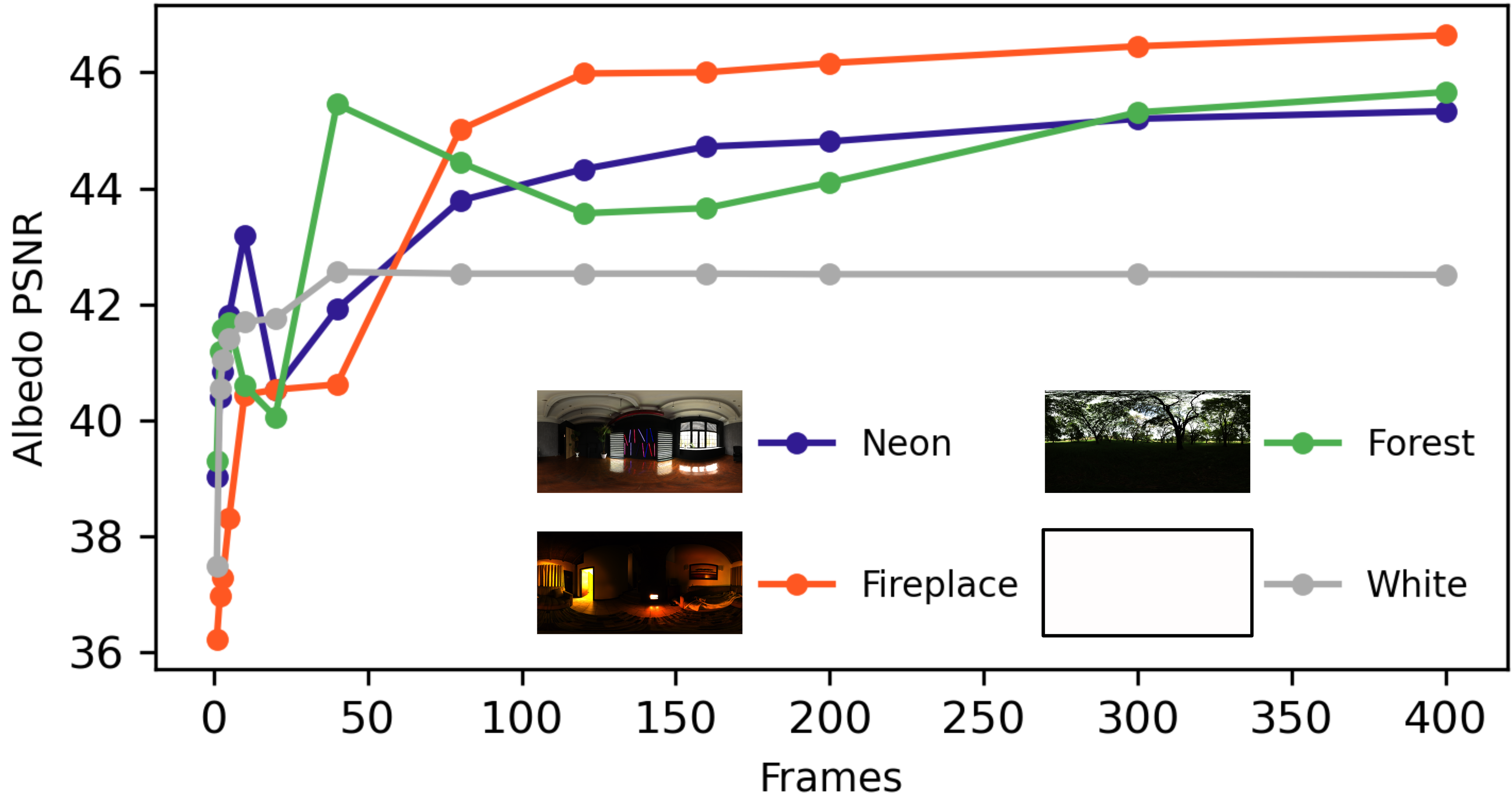}
    \caption{Effect of diversity in hand-held rotations on decomposition. Increasing the diversity of rotated lighting observations improves disentanglement under complex illumination, while uniform white lighting provides no additional constraints.}
    \label{fig:rotation_ablation}
\end{figure}

\section{Additional Analysis}

\noindent\textbf{View Sampling Ablation.}
In Figure~\ref{fig:view_ablation}, we show that material-lighting decomposition gains are independent of view sampling: 25 hand-held views outperform 400 static views. In the main evaluation, we use 400 views for the synthetic static and hand-held settings and 200 views for the turntable setting.

\noindent\textbf{Diversity in Hand-held Rotations.}
Figure~\ref{fig:rotation_ablation} isolates the effect of greater variation in hand-held rotations on albedo-light disentanglement by keeping the camera and geometry fixed and rotating the environment only according to the given trajectory poses. We show the effect under different environment maps with varying lighting complexity. Since the object view is fixed, the disentanglement is based purely on surface-light interactions and improves as we add more frames from the trajectory for diverse lighting conditions while it stagnates under uniform white lighting.

\section{Benchmarking Pose Estimation}
We evaluate our pose estimation pipeline against the state-of-the-art online RGB video reconstruction method FMOV~\cite{shi2025fmov}. FMOV uses a neural surface for both the progressive optimization and global refinement stages. We show in Table~\ref{tab:pose_comparison} that using a better feature matcher and 3D Gaussians---which capture finer geometry---for global refinement helps estimate the pose more precisely. We follow the protocol defined by FMOV and evaluate on 9 sequences from the HO3D~\cite{hampali2020honnotate} dataset. We report the area under the curve (AUC) with a threshold of 10~cm in absolute trajectory error (ATE) and also the relative pose error (RPE), composed of the translation error and the rotation error, respectively.

\begin{table}[t]
    \centering
    \caption{Pose estimation performance on the HO3D dataset. Our 3D Gaussian-based refinement improves both absolute trajectory accuracy and relative pose accuracy over FMOV. *Our implementation.}
    \label{tab:pose_comparison}
    \begin{tabular}{c|ccc}
        \toprule
        Method & $\text{AUC}_\text{ATE}\uparrow$ & $\text{RPE}_t\downarrow$ & $\text{RPE}_r\downarrow$ \\
        \midrule
        FMOV* & 31.19 & 1.03 & 3.69 \\ 
        \midrule
        Ours & \textbf{35.26} & \textbf{0.63} & \textbf{3.00} \\
        \bottomrule
    \end{tabular}
\end{table}

\section{Additional Per-Object Results}

In Tables~\ref{tab:synthetic_capture_configurations_per_object_original} and~\ref{tab:synthetic_capture_configurations_per_object_diffuse}, we provide per-object results for each of the 10 objects and the three capture configurations (static dome~\includegraphics[scale=0.065]{figures/symbols_static_dome.png}, turntable~\includegraphics[scale=0.065]{figures/symbols_turntable.png} and hand-held~\includegraphics[scale=0.065]{figures/symbols_handheld.png}) introduced in our synthetic dataset for a more fine-grained evaluation of the decomposition performance, for both the original and diffuse variants. The proposed hand-held capture setting, either with estimated or ground-truth poses, consistently performs better across all objects, further demonstrating robustness to geometric complexity.

\clearpage

\begin{center}
\small

\begingroup
\captionsetup{type=table}
\refstepcounter{table}
\caption*{\textbf{Table \thetable:} Quantitative per-object results for different capture configurations on the original material variant of our proposed dataset. The best-performing configuration for each object is highlighted in \textbf{bold}, while the second-best is \underline{underlined}.}
\label{tab:synthetic_capture_configurations_per_object_original}
\endgroup

\begin{longtable}{c|c|c|ccc|ccc|c}
    \toprule
    \multirow{2}{*}{Object} & \multirow{2}{*}{Setting} & \multirow{2}{*}{Pose} & \multicolumn{3}{c|}{Albedo} & \multicolumn{3}{c|}{Relighting} & Normal \\
    & & & PSNR$\uparrow$ & SSIM$\uparrow$ & LPIPS$\downarrow$ & PSNR$\uparrow$ & SSIM$\uparrow$ & LPIPS$\downarrow$ & MAE$\downarrow$ \\
    \midrule
    \endfirsthead
    
    \toprule
    \multirow{2}{*}{Object} & \multirow{2}{*}{Setting} & \multirow{2}{*}{Pose} & \multicolumn{3}{c|}{Albedo} & \multicolumn{3}{c|}{Relighting} & Normal \\
    & & & PSNR$\uparrow$ & SSIM$\uparrow$ & LPIPS$\downarrow$ & PSNR$\uparrow$ & SSIM$\uparrow$ & LPIPS$\downarrow$ & MAE$\downarrow$ \\
    \midrule
    \endhead
    
    \bottomrule
    \endfoot
    
    \bottomrule
    \endlastfoot

    \multirow{5}{*}[-2.75ex]{Can} & \includegraphics[scale=0.05]{figures/symbols_static_dome.png} & G.T. & 37.03 & \underline{0.993} & \underline{0.008} & 40.01 & 0.993 & \underline{0.010} & 0.367 \\ 
    \cmidrule{2-10}
    & \multirow{2}{*}[-0.75ex]{\includegraphics[scale=0.05]{figures/symbols_turntable.png}} & Est. & 32.38 & 0.984 & 0.015 & 33.17 & 0.980 & 0.021 & 1.451 \\ 
    \cmidrule{3-10}
    & & G.T. & 38.00 & \underline{0.993} & \underline{0.008} & 35.74 & 0.988 & 0.014 & 1.056 \\
    \cmidrule{2-10}
    & \multirow{2}{*}[-0.75ex]{\includegraphics[scale=0.05]{figures/symbols_handheld.png}} & Est. & \underline{41.07} & \textbf{0.995} & \textbf{0.007} & \underline{40.89} & \underline{0.994} & \textbf{0.008} & \underline{0.325} \\
    \cmidrule{3-10}
    & & G.T. & \textbf{41.22} & \textbf{0.995} & \textbf{0.007} & \textbf{42.10} & \textbf{0.995} & \textbf{0.008} & \textbf{0.260} \\
    \midrule

    \multirow{5}{*}[-2.75ex]{Dino} & \includegraphics[scale=0.05]{figures/symbols_static_dome.png} & G.T. & 44.89 & 0.991 & 0.017 & 39.21 & 0.985 & \underline{0.018} & 0.600 \\ 
    \cmidrule{2-10}
    & \multirow{2}{*}[-0.75ex]{\includegraphics[scale=0.05]{figures/symbols_turntable.png}} & Est. & 45.64 & \underline{0.992} & \underline{0.016} & 37.98 & 0.981 & 0.021 & 0.764 \\ 
    \cmidrule{3-10}
    & & G.T. & 45.56 & \underline{0.992} & 0.019 & 37.92 & 0.980 & 0.022 & 0.782 \\
    \cmidrule{2-10}
    & \multirow{2}{*}[-0.75ex]{\includegraphics[scale=0.05]{figures/symbols_handheld.png}} & Est. & \underline{48.28} & \textbf{0.994} & \textbf{0.013} & \underline{39.68} & \underline{0.986} & \textbf{0.017} & \underline{0.489} \\
    \cmidrule{3-10}
    & & G.T. & \textbf{48.66} & \textbf{0.994} & \textbf{0.013} & \textbf{40.33} & \textbf{0.987} & \textbf{0.017} & \textbf{0.438} \\
    \midrule

    \multirow{5}{*}[-2.75ex]{House} & \includegraphics[scale=0.05]{figures/symbols_static_dome.png} & G.T. & 27.45 & 0.958 & 0.050 & 32.63 & 0.958 & 0.049 & 2.469 \\ 
    \cmidrule{2-10}
    & \multirow{2}{*}[-0.75ex]{\includegraphics[scale=0.05]{figures/symbols_turntable.png}} & Est. & 26.28 & 0.959 & 0.040 & 32.02 & 0.956 & 0.046 & 2.792 \\ 
    \cmidrule{3-10}
    & & G.T. & 26.14 & 0.956 & 0.043 & 31.92 & 0.955 & 0.049 & 2.960 \\
    \cmidrule{2-10}
    & \multirow{2}{*}[-0.75ex]{\includegraphics[scale=0.05]{figures/symbols_handheld.png}} & Est. & \underline{29.20} & \underline{0.965} & \textbf{0.038} & \textbf{34.52} & \textbf{0.969} & \textbf{0.039} & \textbf{2.041} \\
    \cmidrule{3-10}
    & & G.T. & \textbf{29.27} & \textbf{0.966} & \underline{0.039} & \underline{34.30} & \underline{0.968} & \underline{0.041} & \underline{2.056} \\
    \midrule

    \multirow{5}{*}[-2.75ex]{Juice} & \includegraphics[scale=0.05]{figures/symbols_static_dome.png} & G.T. & 34.69 & \underline{0.990} & 0.011 & 36.86 & \underline{0.988} & 0.015 & \underline{0.959} \\ 
    \cmidrule{2-10}
    & \multirow{2}{*}[-0.75ex]{\includegraphics[scale=0.05]{figures/symbols_turntable.png}} & Est. & 36.02 & \underline{0.990} & \underline{0.008} & 36.25 & 0.985 & 0.016 & 1.014 \\ 
    \cmidrule{3-10}
    & & G.T. & 35.53 & 0.988 & 0.011 & 35.90 & 0.983 & 0.019 & 1.021 \\
    \cmidrule{2-10}
    & \multirow{2}{*}[-0.75ex]{\includegraphics[scale=0.05]{figures/symbols_handheld.png}} & Est. & \underline{37.27} & \textbf{0.994} & \textbf{0.006} & \textbf{40.79} & \textbf{0.993} & \textbf{0.009} & \textbf{0.424} \\
    \cmidrule{3-10}
    & & G.T. & \textbf{37.52} & \textbf{0.994} & \textbf{0.006} & \underline{40.73} & \textbf{0.993} & \underline{0.010} & \textbf{0.424} \\
    \midrule

    \multirow{5}{*}[-2.75ex]{Milk} & \includegraphics[scale=0.05]{figures/symbols_static_dome.png} & G.T. & 31.03 & 0.985 & 0.014 & 36.42 & \underline{0.985} & \underline{0.015} & 1.070 \\ 
    \cmidrule{2-10}
    & \multirow{2}{*}[-0.75ex]{\includegraphics[scale=0.05]{figures/symbols_turntable.png}} & Est. & 32.58 & 0.985 & 0.011 & 34.84 & 0.981 & \underline{0.015} & 1.515 \\ 
    \cmidrule{3-10}
    & & G.T. & 31.69 & 0.982 & 0.015 & 34.97 & 0.982 & 0.017 & 1.080 \\
    \cmidrule{2-10}
    & \multirow{2}{*}[-0.75ex]{\includegraphics[scale=0.05]{figures/symbols_handheld.png}} & Est. & \underline{36.92} & \textbf{0.993} & \textbf{0.006} & \textbf{39.16} & \textbf{0.991} & \textbf{0.010} & \underline{0.520} \\
    \cmidrule{3-10}
    & & G.T. & \textbf{36.99} & \underline{0.992} & \underline{0.007} & \underline{39.02} & \textbf{0.991} & \textbf{0.010} & \textbf{0.505} \\
    \midrule

    \multirow{5}{*}[-2.75ex]{Mustard} & \includegraphics[scale=0.05]{figures/symbols_static_dome.png} & G.T. & 41.59 & 0.994 & 0.016 & \underline{41.70} & \textbf{0.994} & \underline{0.014} & \underline{0.314} \\ 
    \cmidrule{2-10}
    & \multirow{2}{*}[-0.75ex]{\includegraphics[scale=0.05]{figures/symbols_turntable.png}} & Est. & 35.20 & 0.983 & 0.031 & 31.22 & 0.976 & 0.033 & 9.921 \\ 
    \cmidrule{3-10}
    & & G.T. & 42.52 & \underline{0.995} & \underline{0.012} & 40.14 & 0.992 & 0.016 & 0.649 \\
    \cmidrule{2-10}
    & \multirow{2}{*}[-0.75ex]{\includegraphics[scale=0.05]{figures/symbols_handheld.png}} & Est. & \underline{43.75} & \textbf{0.996} & \textbf{0.010} & 39.97 & \underline{0.993} & 0.015 & 0.454 \\
    \cmidrule{3-10}
    & & G.T. & \textbf{44.26} & \textbf{0.996} & \textbf{0.010} & \textbf{42.34} & \textbf{0.994} & \textbf{0.013} & \textbf{0.185} \\
    \midrule

    \multirow{5}{*}[-2.75ex]{Ranch} & \includegraphics[scale=0.05]{figures/symbols_static_dome.png} & G.T. & 38.72 & \underline{0.993} & 0.018 & \underline{40.14} & \underline{0.992} & 0.016 & \underline{0.398} \\ 
    \cmidrule{2-10}
    & \multirow{2}{*}[-0.75ex]{\includegraphics[scale=0.05]{figures/symbols_turntable.png}} & Est. & 29.49 & 0.979 & 0.028 & 27.05 & 0.972 & 0.033 & 1.705 \\ 
    \cmidrule{3-10}
    & & G.T. & 36.89 & 0.992 & \underline{0.013} & 36.62 & 0.990 & 0.018 & 0.819 \\
    \cmidrule{2-10}
    & \multirow{2}{*}[-0.75ex]{\includegraphics[scale=0.05]{figures/symbols_handheld.png}} & Est. & \underline{40.74} & \textbf{0.995} & \textbf{0.012} & 36.47 & 0.991 & \underline{0.015} & 0.653 \\
    \cmidrule{3-10}
    & & G.T. & \textbf{40.93} & \textbf{0.995} & \textbf{0.012} & \textbf{41.69} & \textbf{0.994} & \textbf{0.013} & \textbf{0.196} \\
    \midrule

    \multirow{5}{*}[-2.75ex]{Sauce} & \includegraphics[scale=0.05]{figures/symbols_static_dome.png} & G.T. & 45.13 & \underline{0.996} & 0.009 & \underline{42.36} & \underline{0.994} & \underline{0.014} & \underline{0.298} \\ 
    \cmidrule{2-10}
    & \multirow{2}{*}[-0.75ex]{\includegraphics[scale=0.05]{figures/symbols_turntable.png}} & Est. & 36.66 & 0.986 & 0.027 & 28.50 & 0.970 & 0.039 & 2.268 \\ 
    \cmidrule{3-10}
    & & G.T. & 42.89 & \underline{0.996} & \underline{0.007} & 39.24 & 0.991 & 0.018 & 0.914 \\
    \cmidrule{2-10}
    & \multirow{2}{*}[-0.75ex]{\includegraphics[scale=0.05]{figures/symbols_handheld.png}} & Est. & \textbf{49.31} & \textbf{0.998} & \textbf{0.004} & 38.17 & 0.991 & 0.015 & 0.915 \\
    \cmidrule{3-10}
    & & G.T. & \underline{49.25} & \textbf{0.998} & \textbf{0.004} & \textbf{43.10} & \textbf{0.995} & \textbf{0.012} & \textbf{0.179} \\
    \midrule

    \multirow{5}{*}[-2.75ex]{Soup} & \includegraphics[scale=0.05]{figures/symbols_static_dome.png} & G.T. & 31.37 & 0.978 & 0.029 & 32.69 & \underline{0.977} & \underline{0.036} & 1.565 \\ 
    \cmidrule{2-10}
    & \multirow{2}{*}[-0.75ex]{\includegraphics[scale=0.05]{figures/symbols_turntable.png}} & Est. & 32.72 & 0.980 & 0.024 & 31.14 & 0.970 & 0.040 & 2.817 \\ 
    \cmidrule{3-10}
    & & G.T. & 32.29 & \underline{0.982} & \underline{0.022} & 31.87 & 0.976 & 0.037 & 2.358 \\
    \cmidrule{2-10}
    & \multirow{2}{*}[-0.75ex]{\includegraphics[scale=0.05]{figures/symbols_handheld.png}} & Est. & \underline{36.90} & \textbf{0.986} & \textbf{0.021} & \textbf{36.67} & \textbf{0.983} & \textbf{0.030} & \underline{0.959} \\
    \cmidrule{3-10}
    & & G.T. & \textbf{37.18} & \textbf{0.986} & 0.023 & \underline{36.60} & \textbf{0.983} & \textbf{0.030} & \textbf{0.926} \\
    \midrule

    \multirow{5}{*}[-2.75ex]{Waffles} & \includegraphics[scale=0.05]{figures/symbols_static_dome.png} & G.T. & 35.34 & 0.993 & 0.008 & \underline{39.36} & \underline{0.995} & \underline{0.008} & \underline{0.258} \\ 
    \cmidrule{2-10}
    & \multirow{2}{*}[-0.75ex]{\includegraphics[scale=0.05]{figures/symbols_turntable.png}} & Est. & 34.09 & \underline{0.994} & \underline{0.006} & 37.99 & 0.994 & 0.009 & 0.260 \\ 
    \cmidrule{3-10}
    & & G.T. & 29.96 & 0.991 & 0.009 & 35.79 & 0.992 & 0.012 & 0.317 \\
    \cmidrule{2-10}
    & \multirow{2}{*}[-0.75ex]{\includegraphics[scale=0.05]{figures/symbols_handheld.png}} & Est. & \underline{40.23} & 0.993 & 0.009 & 39.12 & 0.991 & 0.011 & 8.824 \\
    \cmidrule{3-10}
    & & G.T. & \textbf{41.90} & \textbf{0.997} & \textbf{0.005} & \textbf{42.16} & \textbf{0.996} & \textbf{0.007} & \textbf{0.132} \\
    \midrule
    \midrule

    \multirow{5}{*}[-2.75ex]{Average} & \includegraphics[scale=0.05]{figures/symbols_static_dome.png} & G.T. & 36.72 & \underline{0.987} & 0.018 & 38.14 & 0.986 & 0.020 & \underline{0.830} \\ 
    \cmidrule{2-10}
    & \multirow{2}{*}[-0.75ex]{\includegraphics[scale=0.05]{figures/symbols_turntable.png}} & Est. & 34.10 & 0.983 & 0.021 & 33.02 & 0.977 & 0.027 & 2.451 \\ 
    \cmidrule{3-10}
    & & G.T. & 36.15 & \underline{0.987} & \underline{0.016} & 36.01 & 0.983 & 0.022 & 1.196 \\
    \cmidrule{2-10}
    & \multirow{2}{*}[-0.75ex]{\includegraphics[scale=0.05]{figures/symbols_handheld.png}} & Est. & \underline{40.37} & \textbf{0.991} & \textbf{0.013} & \underline{38.54} & \underline{0.988} & \underline{0.017} & 1.560 \\
    \cmidrule{3-10}
    & & G.T. & \textbf{40.72} & \textbf{0.991} & \textbf{0.013} & \textbf{40.24} & \textbf{0.989} & \textbf{0.016} & \textbf{0.530}

\end{longtable}
\end{center}

\clearpage

\begingroup
\captionsetup{type=table}
\phantomsection
\makeatletter
\protected@edef\@currentlabel{\thetable}
\makeatother
\caption*{\textbf{Table \thetable:} Quantitative per-object results for different capture configurations on the diffuse material variant of our proposed dataset. The best-performing configuration for each object is highlighted in \textbf{bold}, while the second-best is \underline{underlined}.}
\label{tab:synthetic_capture_configurations_per_object_diffuse}
\endgroup

\begin{center}
\small

\begin{longtable}{c|c|c|ccc|ccc|c}
    \toprule
    \multirow{2}{*}{Object} & \multirow{2}{*}{Setting} & \multirow{2}{*}{Pose} & \multicolumn{3}{c|}{Albedo} & \multicolumn{3}{c|}{Relighting} & Normal \\
    & & & PSNR$\uparrow$ & SSIM$\uparrow$ & LPIPS$\downarrow$ & PSNR$\uparrow$ & SSIM$\uparrow$ & LPIPS$\downarrow$ & MAE$\downarrow$ \\
    \midrule
    \endfirsthead
    
    \toprule
    \multirow{2}{*}{Object} & \multirow{2}{*}{Setting} & \multirow{2}{*}{Pose} & \multicolumn{3}{c|}{Albedo} & \multicolumn{3}{c|}{Relighting} & Normal \\
    & & & PSNR$\uparrow$ & SSIM$\uparrow$ & LPIPS$\downarrow$ & PSNR$\uparrow$ & SSIM$\uparrow$ & LPIPS$\downarrow$ & MAE$\downarrow$ \\
    \midrule
    \endhead
    
    \bottomrule
    \endfoot
    
    \bottomrule
    \endlastfoot

    \multirow{5}{*}[-2.75ex]{Can} & \includegraphics[scale=0.05]{figures/symbols_static_dome.png} & G.T. & 31.38 & 0.987 & 0.011 & 34.86 & 0.986 & \underline{0.015} & 1.076 \\ 
    \cmidrule{2-10}
    & \multirow{2}{*}[-0.75ex]{\includegraphics[scale=0.05]{figures/symbols_turntable.png}} & Est. & 33.11 & 0.985 & 0.012 & 33.77 & 0.982 & 0.017 & 1.575 \\ 
    \cmidrule{3-10}
    & & G.T. & 35.99 & \underline{0.990} & \underline{0.008} & 35.58 & \underline{0.987} & \underline{0.015} & 1.184 \\
    \cmidrule{2-10}
    & \multirow{2}{*}[-0.75ex]{\includegraphics[scale=0.05]{figures/symbols_handheld.png}} & Est. & \underline{40.02} & \textbf{0.995} & \textbf{0.005} & \underline{43.80} & \textbf{0.996} & \textbf{0.007} & \underline{0.342} \\
    \cmidrule{3-10}
    & & G.T. & \textbf{40.40} & \textbf{0.995} & \textbf{0.005} & \textbf{44.62} & \textbf{0.996} & \textbf{0.007} & \textbf{0.287} \\
    \midrule

    \multirow{5}{*}[-2.75ex]{Dino} & \includegraphics[scale=0.05]{figures/symbols_static_dome.png} & G.T. & 41.55 & 0.989 & 0.018 & 37.91 & \underline{0.983} & \underline{0.022} & 1.206 \\ 
    \cmidrule{2-10}
    & \multirow{2}{*}[-0.75ex]{\includegraphics[scale=0.05]{figures/symbols_turntable.png}} & Est. & 44.09 & 0.992 & 0.016 & 36.24 & 0.978 & 0.024 & 1.150 \\ 
    \cmidrule{3-10}
    & & G.T. & 45.74 & 0.993 & \underline{0.015} & 36.89 & 0.981 & 0.024 & 0.974 \\
    \cmidrule{2-10}
    & \multirow{2}{*}[-0.75ex]{\includegraphics[scale=0.05]{figures/symbols_handheld.png}} & Est. & \underline{47.20} & \underline{0.995} & \textbf{0.009} & \underline{40.13} & \textbf{0.990} & \textbf{0.014} & \underline{0.557} \\
    \cmidrule{3-10}
    & & G.T. & \textbf{47.40} & \textbf{0.996} & \textbf{0.009} & \textbf{40.19} & \textbf{0.990} & \textbf{0.014} & \textbf{0.516} \\
    \midrule

    \multirow{5}{*}[-2.75ex]{House} & \includegraphics[scale=0.05]{figures/symbols_static_dome.png} & G.T. & 25.52 & 0.954 & 0.058 & 31.50 & 0.959 & 0.052 & 3.160 \\ 
    \cmidrule{2-10}
    & \multirow{2}{*}[-0.75ex]{\includegraphics[scale=0.05]{figures/symbols_turntable.png}} & Est. & 25.56 & 0.955 & \underline{0.044} & 31.17 & 0.958 & 0.045 & 2.944 \\ 
    \cmidrule{3-10}
    & & G.T. & 25.30 & 0.949 & 0.048 & 31.07 & 0.956 & 0.048 & 3.203 \\
    \cmidrule{2-10}
    & \multirow{2}{*}[-0.75ex]{\includegraphics[scale=0.05]{figures/symbols_handheld.png}} & Est. & \textbf{29.44} & \underline{0.956} & \textbf{0.036} & \textbf{35.38} & \textbf{0.972} & \textbf{0.038} & \underline{2.127} \\
    \cmidrule{3-10}
    & & G.T. & \underline{29.41} & \textbf{0.957} & \textbf{0.036} & \underline{35.03} & \underline{0.971} & \underline{0.040} & \textbf{2.107} \\
    \midrule

    \multirow{5}{*}[-2.75ex]{Juice} & \includegraphics[scale=0.05]{figures/symbols_static_dome.png} & G.T. & 33.41 & 0.988 & 0.012 & 36.87 & \underline{0.986} & 0.016 & 1.187 \\ 
    \cmidrule{2-10}
    & \multirow{2}{*}[-0.75ex]{\includegraphics[scale=0.05]{figures/symbols_turntable.png}} & Est. & 32.98 & \underline{0.989} & \underline{0.009} & 36.53 & \underline{0.986} & \underline{0.015} & 1.002 \\ 
    \cmidrule{3-10}
    & & G.T. & 32.42 & 0.987 & 0.012 & 36.47 & 0.985 & 0.016 & 1.053 \\
    \cmidrule{2-10}
    & \multirow{2}{*}[-0.75ex]{\includegraphics[scale=0.05]{figures/symbols_handheld.png}} & Est. & \underline{36.38} & \textbf{0.993} & \textbf{0.006} & \textbf{41.51} & \textbf{0.993} & \textbf{0.009} & \textbf{0.437} \\
    \cmidrule{3-10}
    & & G.T. & \textbf{36.52} & \textbf{0.993} & \textbf{0.006} & \underline{41.38} & \textbf{0.993} & \textbf{0.009} & \underline{0.443} \\
    \midrule

    \multirow{5}{*}[-2.75ex]{Milk} & \includegraphics[scale=0.05]{figures/symbols_static_dome.png} & G.T. & 29.14 & \underline{0.984} & 0.016 & 34.06 & 0.978 & \underline{0.018} & 2.153 \\ 
    \cmidrule{2-10}
    & \multirow{2}{*}[-0.75ex]{\includegraphics[scale=0.05]{figures/symbols_turntable.png}} & Est. & 30.38 & 0.981 & 0.014 & 31.84 & 0.973 & 0.019 & 2.566 \\ 
    \cmidrule{3-10}
    & & G.T. & 30.87 & 0.979 & 0.016 & 31.80 & 0.972 & 0.020 & 2.675 \\
    \cmidrule{2-10}
    & \multirow{2}{*}[-0.75ex]{\includegraphics[scale=0.05]{figures/symbols_handheld.png}} & Est. & \textbf{36.71} & \textbf{0.987} & \textbf{0.004} & \textbf{39.81} & \textbf{0.990} & \textbf{0.010} & \underline{0.715} \\
    \cmidrule{3-10}
    & & G.T. & \underline{36.36} & \textbf{0.987} & \underline{0.005} & \underline{39.63} & \underline{0.989} & \textbf{0.010} & \textbf{0.681} \\
    \midrule

    \multirow{5}{*}[-2.75ex]{Mustard} & \includegraphics[scale=0.05]{figures/symbols_static_dome.png} & G.T. & 34.85 & 0.990 & 0.019 & 35.80 & 0.988 & 0.020 & 1.047 \\ 
    \cmidrule{2-10}
    & \multirow{2}{*}[-0.75ex]{\includegraphics[scale=0.05]{figures/symbols_turntable.png}} & Est. & 31.96 & 0.978 & 0.044 & 29.47 & 0.971 & 0.038 & 2.200 \\ 
    \cmidrule{3-10}
    & & G.T. & 40.24 & \underline{0.995} & \underline{0.012} & 36.70 & 0.989 & 0.017 & 0.653 \\
    \cmidrule{2-10}
    & \multirow{2}{*}[-0.75ex]{\includegraphics[scale=0.05]{figures/symbols_handheld.png}} & Est. & \underline{43.00} & \textbf{0.997} & \textbf{0.009} & \underline{39.61} & \underline{0.992} & \underline{0.015} & \underline{0.556} \\
    \cmidrule{3-10}
    & & G.T. & \textbf{43.94} & \textbf{0.997} & \textbf{0.009} & \textbf{41.91} & \textbf{0.993} & \textbf{0.013} & \textbf{0.248} \\
    \midrule

    \multirow{5}{*}[-2.75ex]{Ranch} & \includegraphics[scale=0.05]{figures/symbols_static_dome.png} & G.T. & 32.84 & 0.991 & 0.017 & 33.91 & 0.988 & 0.019 & 1.335 \\ 
    \cmidrule{2-10}
    & \multirow{2}{*}[-0.75ex]{\includegraphics[scale=0.05]{figures/symbols_turntable.png}} & Est. & 32.61 & 0.988 & 0.021 & 31.84 & 0.982 & 0.027 & 1.393 \\ 
    \cmidrule{3-10}
    & & G.T. & 42.33 & \textbf{0.995} & \underline{0.005} & 38.08 & 0.992 & 0.017 & 0.770 \\
    \cmidrule{2-10}
    & \multirow{2}{*}[-0.75ex]{\includegraphics[scale=0.05]{figures/symbols_handheld.png}} & Est. & \underline{42.90} & \underline{0.994} & \textbf{0.004} & \underline{39.46} & \underline{0.994} & \underline{0.013} & \underline{0.635} \\
    \cmidrule{3-10}
    & & G.T. & \textbf{43.20} & \underline{0.994} & \textbf{0.004} & \textbf{43.78} & \textbf{0.995} & \textbf{0.012} & \textbf{0.227} \\
    \midrule

    \multirow{5}{*}[-2.75ex]{Sauce} & \includegraphics[scale=0.05]{figures/symbols_static_dome.png} & G.T. & 36.68 & 0.990 & 0.012 & 36.97 & 0.990 & 0.019 & 0.973 \\ 
    \cmidrule{2-10}
    & \multirow{2}{*}[-0.75ex]{\includegraphics[scale=0.05]{figures/symbols_turntable.png}} & Est. & 37.62 & 0.989 & 0.023 & 32.27 & 0.976 & 0.037 & 2.208 \\ 
    \cmidrule{3-10}
    & & G.T. & 46.02 & \underline{0.997} & \underline{0.006} & 37.98 & 0.991 & 0.017 & \underline{0.569} \\
    \cmidrule{2-10}
    & \multirow{2}{*}[-0.75ex]{\includegraphics[scale=0.05]{figures/symbols_handheld.png}} & Est. & \underline{48.55} & \textbf{0.998} & \textbf{0.003} & \underline{40.03} & \underline{0.994} & \underline{0.012} & 0.891 \\
    \cmidrule{3-10}
    & & G.T. & \textbf{48.69} & \textbf{0.998} & \textbf{0.003} & \textbf{42.71} & \textbf{0.995} & \textbf{0.011} & \textbf{0.225} \\
    \midrule

    \multirow{5}{*}[-2.75ex]{Soup} & \includegraphics[scale=0.05]{figures/symbols_static_dome.png} & G.T. & 27.61 & 0.967 & 0.038 & 30.03 & 0.964 & 0.046 & 3.006 \\ 
    \cmidrule{2-10}
    & \multirow{2}{*}[-0.75ex]{\includegraphics[scale=0.05]{figures/symbols_turntable.png}} & Est. & 29.56 & 0.966 & 0.030 & 31.58 & 0.963 & 0.040 & 2.922 \\ 
    \cmidrule{3-10}
    & & G.T. & 29.45 & \underline{0.968} & 0.027 & 32.06 & 0.967 & 0.040 & 2.664 \\
    \cmidrule{2-10}
    & \multirow{2}{*}[-0.75ex]{\includegraphics[scale=0.05]{figures/symbols_handheld.png}} & Est. & \underline{36.98} & \textbf{0.983} & \textbf{0.017} & \textbf{38.82} & \textbf{0.987} & \textbf{0.025} & \underline{1.038} \\
    \cmidrule{3-10}
    & & G.T. & \textbf{37.23} & \textbf{0.983} & \underline{0.018} & \underline{38.66} & \underline{0.986} & \underline{0.026} & \textbf{0.943} \\
    \midrule

    \multirow{5}{*}[-2.75ex]{Waffles} & \includegraphics[scale=0.05]{figures/symbols_static_dome.png} & G.T. & 33.82 & 0.990 & 0.009 & 36.57 & 0.990 & 0.013 & 0.748 \\ 
    \cmidrule{2-10}
    & \multirow{2}{*}[-0.75ex]{\includegraphics[scale=0.05]{figures/symbols_turntable.png}} & Est. & 31.87 & \underline{0.993} & \underline{0.004} & 35.69 & \underline{0.992} & \underline{0.010} & 0.330 \\ 
    \cmidrule{3-10}
    & & G.T. & 28.13 & 0.989 & 0.006 & 34.37 & 0.991 & 0.011 & 0.449 \\
    \cmidrule{2-10}
    & \multirow{2}{*}[-0.75ex]{\includegraphics[scale=0.05]{figures/symbols_handheld.png}} & Est. & \underline{42.11} & \textbf{0.998} & \textbf{0.003} & \textbf{45.15} & \textbf{0.997} & \textbf{0.006} & \underline{0.149} \\
    \cmidrule{3-10}
    & & G.T. & \textbf{42.30} & \textbf{0.998} & \textbf{0.003} & \underline{45.01} & \textbf{0.997} & \textbf{0.006} & \textbf{0.138} \\
    \midrule
    \midrule

    \multirow{5}{*}[-2.75ex]{Average} & \includegraphics[scale=0.05]{figures/symbols_static_dome.png} & G.T. & 32.68 & 0.983 & 0.021 & 34.85 & 0.981 & 0.024 & 1.589 \\ 
    \cmidrule{2-10}
    & \multirow{2}{*}[-0.75ex]{\includegraphics[scale=0.05]{figures/symbols_turntable.png}} & Est. & 32.97 & 0.982 & 0.022 & 33.04 & 0.976 & 0.027 & 1.829 \\ 
    \cmidrule{3-10}
    & & G.T. & 35.65 & \underline{0.984} & \underline{0.016} & 35.10 & 0.981 & \underline{0.023} & 1.419 \\
    \cmidrule{2-10}
    & \multirow{2}{*}[-0.75ex]{\includegraphics[scale=0.05]{figures/symbols_handheld.png}} & Est. & \underline{40.33} & \textbf{0.990} & \textbf{0.010} & \underline{40.37} & \underline{0.990} & \textbf{0.015} & \underline{0.745} \\
    \cmidrule{3-10}
    & & G.T. & \textbf{40.55} & \textbf{0.990} & \textbf{0.010} & \textbf{41.29} & \textbf{0.991} & \textbf{0.015} & \textbf{0.582}

\end{longtable}
\end{center}

\end{document}